%% file: main.tex
\documentclass[10pt,twocolumn,letterpaper]{article}

\usepackage{cvpr}

\input{preamble}
\usepackage{makecell}
\usepackage{graphicx}        
\usepackage{booktabs}        
\usepackage{amsmath}         
\usepackage{caption}         
\usepackage{adjustbox}       
\usepackage[table]{xcolor}   
\definecolor{navy}{RGB}{0, 0, 128} 
\definecolor{maroon}{RGB}{128, 0, 0} 
\usepackage{subcaption}
\definecolor{cvprblue}{rgb}{0.21,0.49,0.74}
\usepackage[pagebackref,breaklinks,colorlinks,allcolors=cvprblue]{hyperref}

\makeatletter
\renewcommand*{\@fnsymbol}[1]{\ensuremath{\ifcase#1\or *\or \dagger\or \ddagger\or
    \mathsection\or \mathparagraph\or \|\or **\or \dagger\dagger
    \or \ddagger\ddagger \else\@ctrerr\fi}}
\makeatother

\title{ProOOD: Prototype-Guided Out-of-Distribution 3D Occupancy Prediction}

\author{Yuheng Zhang$^{1}$ \qquad Mengfei Duan$^{1}$ \qquad Kunyu Peng$^{2,3,}$\thanks{Corresponding authors (e-mail: {\tt kailun.yang@hnu.edu.cn, kunyu.peng@kit.edu}).} \qquad Yuhang Wang$^{1}$ \qquad Di Wen$^{2}$\\Danda Pani Paudel$^{3}$ \qquad Luc Van Gool$^{3}$ \qquad Kailun Yang$^{1,*}$
\\
\normalsize 
$^{1}$Hunan University 
\normalsize \quad 
$^{2}$Karlsruhe Institute of Technology
\normalsize \quad 
$^{3}$INSAIT, Sofia University ``St. Kliment Ohridski''
}

\let\oldtwocolumn\twocolumn
\renewcommand\twocolumn[1][]{%
    \oldtwocolumn[{#1}{
    \begin{center}
    \vskip -6ex
        \centering
        \includegraphics[width=.95\textwidth]{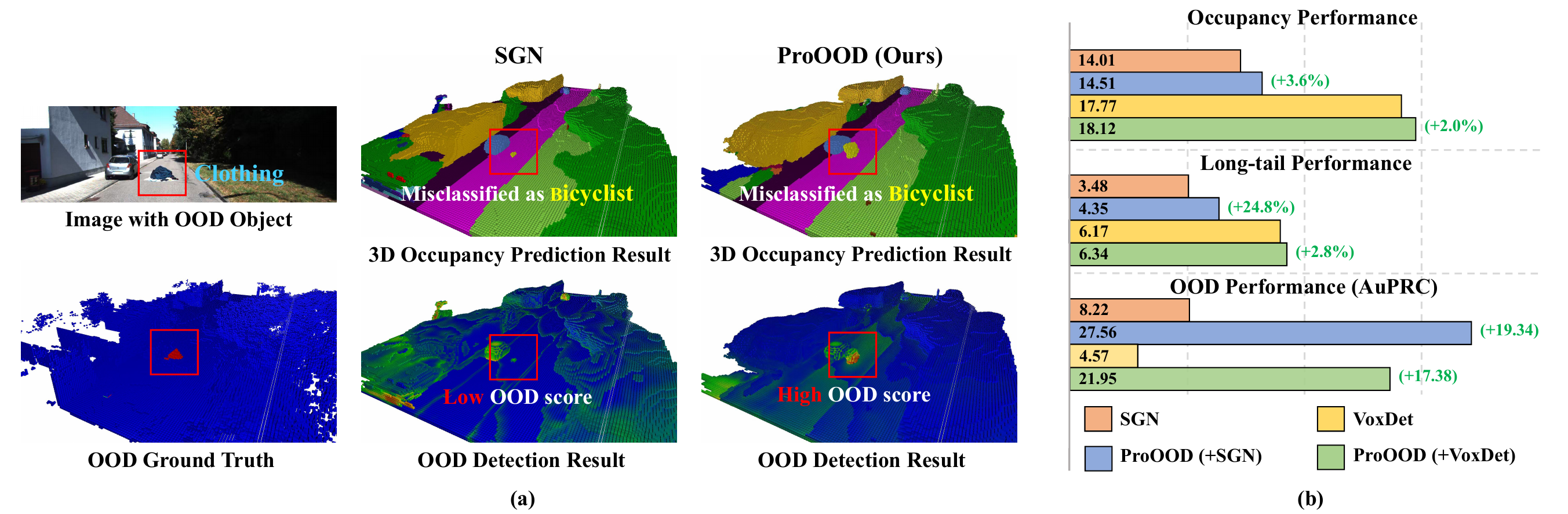}
        \vskip -2ex
        \captionof{figure} {{\textbf{(a) Qualitative results on SemanticKITTI.} The top row displays semantic occupancy predictions. The bottom row displays OOD detection maps, where OOD scores are visualized from low (\textcolor{blue}{blue}) to high (\textcolor{red}{red}).  
        \textbf{(b) Performance comparisons with baselines.} For SemanticKITTI, we report occupancy mIoU in the top panel and long-tailed classes mIoU in the middle panel. For VAA-KITTI, we report OOD detection performance using $AuPRC_r$ in the bottom panel.
        }}
        \label{fig:comparsion}
    \end{center}
    }]
}

\begin{document}
\maketitle
\input{sec/0_abstract}    
\input{sec/1_intro}
\input{sec/2_relatedWork}
\input{sec/3_methodology}
\input{sec/4_experiments}
\input{sec/5_conclusion}
\clearpage

\input{sec/ack}
{
    \small
    \bibliographystyle{ieeenat_fullname}
    \bibliography{main}
}

\input{sec/X_suppl}

\end{document}

%% file: preamble.tex
\definecolor{codegreen}{rgb}{0.0,0.6,0.0}
\definecolor{mygray}{gray}{.9}
\definecolor{mygreen}{RGB}{0, 176, 80}

\usepackage{pifont}
\usepackage{multirow}



%% file: sec/0_abstract.tex
\begin{abstract}
3D semantic occupancy prediction is central to autonomous driving, yet current methods are vulnerable to long-tailed class bias and out-of-distribution (OOD) inputs, often overconfidently assigning anomalies to rare classes. We present ProOOD, a lightweight, plug-and-play method that couples prototype-guided refinement with training-free OOD scoring. ProOOD comprises (i) prototype-guided semantic imputation that fills occluded regions with class-consistent features, (ii) prototype-guided tail mining that strengthens rare-class representations to curb OOD absorption, and (iii) EchoOOD, which fuses local logit coherence with local and global prototype matching to produce reliable voxel-level OOD scores. Extensive experiments on five datasets demonstrate that ProOOD achieves state-of-the-art performance on both in-distribution 3D occupancy prediction and OOD detection. On SemanticKITTI, it surpasses baselines by +3.57\% mIoU overall and +24.80\% tail-class mIoU; on VAA-KITTI, it improves AuPRC$_r$ by +19.34 points, with consistent gains across benchmarks. These improvements yield more calibrated occupancy estimates and more reliable OOD detection in safety-critical urban driving. The source code is publicly available at \url{https://github.com/7uHeng/ProOOD}.
\end{abstract}

%% file: sec/1_intro.tex
\section{Introduction}

Accurate semantic perception is vital for autonomous driving, enabling reliable understanding of complex environments for safe planning and navigation~\cite{hu2023planning,tian2023occ3d}. 
However, point cloud and Bird’s-Eye-View (BEV) methods, while effective for object localization, are constrained by sparse or top-down representations, limiting dense geometric reasoning~\cite{peng2023bevsegformer,pan2023baeformer,li2023bev,li2022bevformer}. 
To overcome this, 3D semantic occupancy prediction has emerged as a promising paradigm that generates voxel-level, geometry-aware semantic maps for precise spatial reasoning and safer motion planning~\cite{cao2022monoscene,wei2023surroundocc}. Recent camera-based approaches~\cite{tang2024sparseocc,li2022bevformer,ma2024cotr,jiang2024symphonize,li2023voxformer,mei2024camera} further improve joint geometric-semantic modeling, advancing scene comprehension and planning reliability.
However, autonomous vehicles operate in complex real-world environments where unknown objects and anomalous scenarios are prevalent, including construction barriers, animals, and extreme weather conditions~\cite{kong2025_3d_4d_world,tian2024tokenize,liu2024surroundsdf,feng2025rap,alama2025rayfronts,liu2025omniscene,deng2024opengraph,xu2025waymo_open}.
These situations often lie outside the training distribution, posing severe challenges to model reliability. 
Out-of-distribution detection for 3D occupancy prediction (OccOoD) is crucial for reliable scene understanding in safety-critical applications~\cite{zhang2025occood}. 
It enables models to identify and handle novel environments, enhancing robustness and trustworthiness in real-world deployment. 
Unlike BEV-based OOD detection, OccOoD operates in full 3D space, capturing volumetric uncertainties that BEV projections overlook, which is vital for accurate obstacle perception and safe navigation and autonomous.

Zhang~\textit{et al.}~\cite{zhang2025occood} pioneered OccOoD benchmarks, yet their method suffers from overconfident misclassification under distribution shifts, where even trivial anomalies (\eg, \textit{clothing}) are falsely recognized as in-distribution classes (\eg, \textit{bicyclist}), as shown in Fig.~\ref{fig:comparsion}(a). This reveals a key limitation: current models lack principled mechanisms to jointly model long-tail calibration and voxel-level uncertainty, both essential for robust 3D OOD detection.
In practice, existing models tend to show \textit{strong classification}, forcefully assigning anomalous objects to rare known categories due to overconfidence in dominant classes and poor calibration for infrequent ones.
As shown in Fig.~\ref{fig:comparsion}(b), this overconfidence results in degraded OOD detection performance despite high in-distribution accuracy.
This stems from two interrelated issues. First, models trained on long-tailed datasets such as SemanticKITTI~\cite{behley2019semantickitti}, where rare classes constitute less than two percent of samples, tend to overassign ambiguous regions to dominant categories due to poor calibration on tail classes. 
Second, while existing OOD scoring methods, \eg, maximum softmax~\cite{xia2022augmenting,liu2023gen}, entropy~\cite{zhang2018generalized,wu2022entropy}, and energy~\cite{choi2023balanced,wang2021energy} are tailored for per-class classification tasks, recent studies~\cite{zhang2025occood} indicate limited performance when applied directly to voxel-level prediction tasks, due to the absence of mechanisms for capturing spatial structure and semantic context in 3D scenes. 

This technical limitation reflects a deeper issue: the model lacks adequate mechanisms to represent tail-class semantics and express uncertainty towards unseen objects, which are both essential for the OccOoD task. 
Therefore, OOD detection and long-tail learning are inherently coupled in 3D occupancy prediction. Better recognition of tail classes improves robustness to rare in-distribution objects and enhances OOD sensitivity by reducing prediction overconfidence and improving model calibration.

Although recent studies have examined OOD detection under sensor corruptions in OCCUQ~\cite{heidrich2025occuq} and proposed semantic-aware anomaly scoring for post-hoc detection in OccOoD~\cite{zhang2025occood}, these efforts have not sufficiently explored the influence of long-tail class distributions on OOD detection performance within 3D occupancy prediction. More importantly, they lack a unified framework explicitly linking long-tail learning with uncertainty quantification in the occupied voxel space. Without such a principled connection and joint optimization, existing methods remain limited in achieving robust and generalizable OccOoD performance across diverse and imbalanced real-world scenarios.

To address these challenges, we propose ProOOD, a prototype-guided framework that unifies long-tail learning and OOD detection for 3D occupancy prediction. By leveraging class-wise prototypes, ProOOD enhances semantic completion, strengthens tail-class representation, and quantifies voxel-wise uncertainty, achieving robust and generalizable OccOoD performance.
Our approach consists of three key components:
First, depth-based query methods~\cite{li2023voxformer,yu2024context} often rely on local geometric consistency for occlusion completion, which fails to ensure semantic plausibility. 
We introduce Prototype-Guided Semantic Imputation (PGSI) to improve completion quality by aligning filled regions with class-level semantic prototypes.
Second, we mitigate the \textit{strong classification} by introducing the Prototype-Guided Tail Mining (PGTM) module, thereby enhancing the model's modeling capability for rare categories and improving performance on both occupancy prediction and OOD detection.
Third, we propose EchoOOD, a prototype-based scoring mechanism that enforces semantic consistency via three cues: local logit alignment, local prototype matching, and global prototype matching, enabling effective OOD detection.
Finally, the ProOOD solution is consistently beneficial atop mainstream voxel-based frameworks, \eg, SGN~\cite{mei2024camera} and VoxDet~\cite{li2025voxdet} (see Fig.~\ref{fig:comparsion}(b)). 

Extensive experiments on five benchmarks show that ProOOD achieves state-of-the-art performance on both in-distribution 3D occupancy prediction and out-of-distribution detection, generalizing to real-world urban driving scenes. 
On SemanticKITTI~\cite{behley2019semantickitti}, it surpasses the baseline by $3.57\%$ mIoU overall and achieves a $24.80\%$ IoU gain on tail classes. 
On VAA-KITTI~\cite{zhang2025occood}, these improvements translate to a $19.34$ points gain in AuPRC$_r$, reflecting enhanced reliability for safety-critical perception tasks.
Our contributions are summarized as follows:

\begin{itemize}
\item We present ProOOD, the first to address OccOoD from a voxel prototype-guided perspective that jointly improves 3D occupancy prediction and OOD detection.
\item We introduce Prototype-Guided Semantic Imputation and Prototype-Guided Tail Mining to refine semantic completion and strengthen rare-class representation.
\item We design EchoOOD, a prototype-based scoring strategy enforcing both local and global semantic consistency for reliable voxel-level OOD detection.
\item Extensive experiments across multiple benchmarks show that ProOOD achieves state-of-the-art in-distribution and OOD performance, yielding significant gains on tail classes and effective generalization to real-world scenes.
\end{itemize}

%% file: sec/2_relatedWork.tex
\section{Related Work}
\label{sec:relatedWork}
\textbf{3D Occupancy Prediction.}
3D occupancy prediction aims to recover the occupancy state and semantic label of each voxel from sensor observations. 
Recent advances in 3D representation~\cite{li2025voxdet,ma2024cotr,yang2025csoocc,bai2024cvformer,huang2024gaussianformer,zuo2025quadricformer,huang2025gaussianformer2}, context modeling~\cite{chen2025alocc,li2024hierarchical_context,liu2025disentangling}, modality fusion~\cite{zhang2024occfusion,li2024occmamba,zhang2025occloff,chu2024afocc,zhao2025gaussianformer3d}, and generative learning~\cite{gao2025loc,tan2023ovo,wang2024occgen,zhu2024nucraft,wang2025diffusion_generative} have collectively improved both estimation quality and efficiency. 
Prototype-based queries are explored~\cite{kim2024protoocc,oh2025_3d_prototype}, but remain agnostic to the long-tail distribution challenge. 
Meanwhile, recent works on open-vocabulary perception~\cite{zheng2024veon,jiang2024openocc,zhang2025clip_occ,boeder2025langocc,yu2024language,li2025ago,li2025pgocc,vobecky2024pop_3d} and occupancy world models~\cite{gu2024dome,jin2025occtens,li2024syntheocc,li2025semi_world} extend semantic understanding beyond closed sets and mitigate long-tail issues, but remain ineffective under OOD conditions, failing to generalize beyond training semantic space. 
OccOoD~\cite{zhang2025occood} and OCCUQ~\cite{heidrich2025occuq} introduce OOD awareness into occupancy prediction, enabling uncertainty estimation~\cite{wang2024reliocc,su2024alpha} and unseen semantics detection in 3D scenes. 
Yet, OccOoD lacks feature-level OOD detection, whereas OCCUQ mainly targets sensor-failure OOD cases.
Although several works~\cite{yu2025shtocc,wang2025occ_exoskeleton,lahoud2024long_tail_3D,xu2025cigocc} reveal the challenges of long-tail categories through class re-balancing or feature enhancement, they still struggle to maintain robustness across OOD and long-tail scenarios.  
In real-world applications, rare or unseen categories are often misidentified, posing potential safety risks. 
In contrast, our method leverages prototype learning to model representations of long-tail categories, enabling more robust and comprehensive scene understanding in real-world driving. 

\noindent\textbf{Out-of-Distribution Segmentation.}
OOD segmentation aims to identify anomalies in scenes while maintaining accurate recognition of in-distribution classes~\cite{laskar2025dataset_unknown,bogdoll2024anovox,batten2025improving_weather}. 
Most existing methods~\cite{grcic2023advantages,rai2024mask2anomaly,delic2024outlier} focus on 2D road-scene images, using per-pixel or mask-based transformer architectures combined with uncertainty estimation~\cite{mukhoti2018evaluating,corbiere2019addressing,hendrycks2019scaling,jung2021standardized,cen2021deep}, reconstruction~\cite{lis2019detecting,xia2020synthesize,haldimann2019not,vojir2021road,grcic2021dense}, or outlier exposure strategies~\cite{chan2021entropy,tian2022pixel,grcic2022densehybrid,liu2023residual,choi2023balanced}. 
Recent efforts have begun extending OOD detection to LiDAR data~\cite{li2025relative,kosel2024revisiting,miandashti2025ood_lidar,shojaei2024uncertainty,soum2025hd_ood3d,huang2022ood_lidar,faulkner2025finding}, yet few consider its interaction with long-tailed class distributions.
In 3D scenes, long-tailed class distributions further complicate OOD detection: rare classes have limited training samples, which can cause semantic ambiguity and lead the model to misclassify anomalies as tail instances or overlook unseen objects. 
Recent studies~\cite{lahoud2024long_tail_3D,yu2025shtocc,peri2023towards_long_tailed,
meng2023learning_rich_coarse,jiang2022improving,fan2025mgaf} explicitly address the long-tail problem to improve tail-class recognition. 
Some approaches~\cite{heidrich2025occuq,zhang2025occood} integrate OOD detection into 3D tasks, leveraging uncertainty quantification and post-hoc detection 
to identify unseen classes and anomalies.
However, effectively combining long-tail learning with OOD detection in 3D remains underexplored. 
Our method explicitly strengthens tail-class representations, thereby enhancing both semantic and overall OOD understanding performance.

%% file: sec/3_methodology.tex
\begin{figure*}[!t]
    \centering
    \includegraphics[width=1\linewidth]{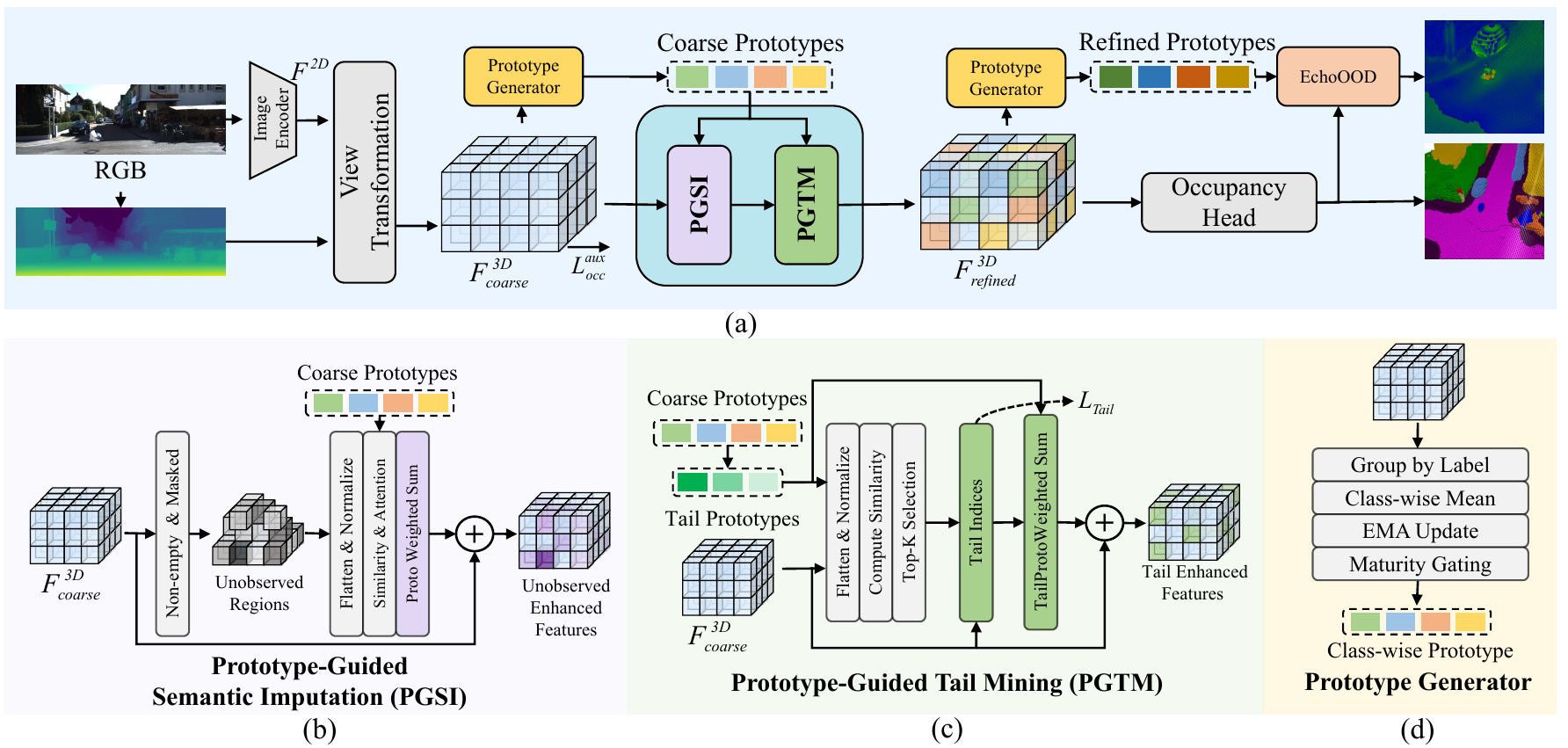}
        \vspace{-5ex}
    \caption{Overview of the proposed ProOOD framework. 
    Coarse 3D features are derived from a 2D backbone and a view transformation module. These features are then refined by PGSI and PGTM modules, which leverage class-wise prototypes updated via EMA to enhance semantic completion and improve tail-class sensitivity. The resulting refined features are used to update the class prototypes, which in turn support both semantic occupancy prediction and OOD detection through the EchoOOD mechanism.}
        \vspace{-3ex}
    \label{fig:pgocc}
\end{figure*}

\section{Methodology}
\label{sec:method}

\subsection{Preliminary}

\noindent\textbf{Task and notation.}
Given single or multi-view images, we estimate a voxel grid of size $(H,W,Z)$.
Let $\mathbf{F}\in\mathbb{R}^{H\times W\times Z\times C}$ denote the 3D feature volume and 
$\mathbf{L}\in\mathbb{R}^{H\times W\times Z\times K_{\mathrm{cls}}}$ represent the class logits for $K_{\mathrm{cls}}$ semantic classes.
An occupancy head predicts semantic results $\hat{\mathbf{Y}}\in\{0,\dots,K_{\mathrm{cls}}\}^{H\times W\times Z}$.
An OOD scoring function $f$ yields a voxel-wise anomaly map $\mathbf{A}=f(\mathbf{F},\mathbf{L})\in[0,1]^{H\times W\times Z}$ without extra supervision or training.

\subsection{Method Overview}
\label{subsec:overview}
3D occupancy prediction aims to reconstruct a complete 3D scene representation from sparse and partial observations. This task is inherently challenging due to occlusions and the structural complexity of real-world scenes~\cite{tian2023occ3d}. 
Current methods~\cite{mei2024camera,yu2024context} over-rely on geometric consistency to fill occluded regions, under-modeling long-tail classes and lacking effective OOD voxel sensitivity.
To address these, we propose \emph{ProOOD}, a plug-and-play method that leverages class-wise voxel prototypes (Sec.~\ref{subsec:prototype_learning}) to guide semantic completion in unobserved regions while simultaneously enhancing sensitivity to tail classes and out-of-distribution objects, all without modifying the backbone. Unlike methods that use prototypes as query tokens~\cite{kim2024protoocc,oh2025_3d_prototype}, ProOOD actively refines intermediate features through prototype-guided modulation, enabling consistent improvement in both long-tail performance and OOD awareness. 

As shown in Fig.~\ref{fig:pgocc}, the pipeline begins with a 2D image encoder that extracts multi-scale features $F^{2D}$, which are projected into an initial 3D volume $F^{3D}$ through the view transformation module. 
A depth-guided proposal then generates coarse 3D features $F^{3D}_{\text{coarse}}$. The \emph{Prototype-Guided Semantic Imputation (PGSI)} module refines $F^{3D}_{\text{coarse}}$ by filling occluded areas with semantically consistent content. The \emph{Prototype-Guided Tail Mining (PGTM)} module subsequently follows, identifying and reinforcing voxels likely belonging to underrepresented classes to improve calibration and reduce overconfidence in dominant categories (Sec.~\ref{subsec:proto_refine}). 
The enhanced features are then passed to the original 3D backbone to produce refined representations $F^{3D}_{\text{refined}}$. 
From these, semantic occupancy predictions are derived via the occupancy head, and OOD scores are computed by the \emph{EchoOOD} mechanism (Sec.~\ref{subsec:ood_scoring}).

\subsection{Class-wise Voxel Prototype Learning}
\label{subsec:prototype_learning}

Voxel features in 3D scenes tend to form compact, well-separated clusters by semantic class, which motivates a prototype-based formulation. 
By learning representative prototypes per class, our method enforces semantic consistency across the occupancy field, improving spatial reasoning and scene understanding.
For each non-empty class $k\in\{1,\dots,K_{\mathrm{cls}}\}$, we maintain a global exponential moving–average (EMA) prototype $\mathbf{p}^{\mathrm{g}}_k\in\mathbb{R}^C$ that tracks the canonical representation of that class over training.
Prototypes are updated \emph{only} with ground-truth voxels to ensure semantic purity, using refined voxel features $\mathbf{x}_i^{(t)}$ extracted from $F^{3D}_{\text{refined}}$ at iteration $t$; empty voxels (class $0$) are excluded.
Let $\Omega^{\mathrm{gt}}_k{}^{(t)}=\{\,i\mid y_i=k\ \text{in the current batch}\,\}$, where $y_i$ denotes the ground-truth label.
With EMA momentum $\beta\in(0,1)$, indicator $\mathbf{1}\{\cdot\}$, and stop-gradient operator $\operatorname{sg}(\cdot)$, the update is:
\begin{equation}
\label{eq:ema_proto}
\begin{aligned}
\mathbf{p}^{\mathrm{g}}_k{}^{(t)} &= \mathbf{p}^{\mathrm{g}}_k{}^{(t-1)}
+ \gamma_k^{(t)}\!\left(\mathbf{m}_k^{(t)} - \mathbf{p}^{\mathrm{g}}_k{}^{(t-1)}\right),\\
\gamma_k^{(t)} &= \beta\,\mathbf{1}\!\left\{\,\big|\Omega^{\mathrm{gt}}_k{}^{(t)}\big|>0\,\right\},\\
\mathbf{m}_k^{(t)} &= \frac{1}{\big|\Omega^{\mathrm{gt}}_k{}^{(t)}\big|}\sum_{i\in\Omega^{\mathrm{gt}}_k{}^{(t)}}
\operatorname{sg}\!\big(\mathbf{x}_i^{(t)}\big).
\end{aligned}
\end{equation}
This design leverages supervised evidence while remaining aligned with the evolving embedding space.
If $\big|\Omega^{\mathrm{gt}}_k{}^{(t)}\big|=0$, the update is skipped.
The stop-gradient prevents backpropagation from updating prototypes and avoids destabilizing the feature encoder.

\noindent\textbf{Maturity gating.}
Prototypes of poor quality fail to effectively distinguish between categories, limiting the model’s ability to model long-tail classes; to prevent such immature prototypes from influencing downstream modules, we gate their usage while maintaining continuous EMA updates throughout training.
Let $\sigma_k^{2,(t)}=\mathrm{Var}\big(\{\mathbf{x}_i^{(t)}\mid y_i=k\}\big)$ denote the current-batch intra-class variance and define a quality score $q_k^{(t)}=\frac{1}{\sigma_k^{2,(t)}+\epsilon}$ with a small $\epsilon>0$.
A class-$k$ prototype is consulted only if $t\ge t_{\mathrm{warm}}$, $q_k^{(t)}>\theta_t$, and $\big|\Omega^{\mathrm{gt}}_k{}^{(t)}\big|\ge n_{\min}$, where
\begin{equation}
\label{eq:theta}
\theta_t = \theta_{\max} \quad \text{for } t \geq t_{\mathrm{warm}},
\end{equation}
and $\theta_{\max}$ is a positive scalar calibrated to the empirical range of $q_k^{(t)}$; $n_{\min}\in\mathbb{N}$ is a small minimum-count threshold (\textit{e.g}., $n_{\min}=2$) to avoid unreliable statistics.
This gate applies to the prototype usage in PGSI and PGTM (Sec.~\ref{subsec:proto_refine}) as well as to the global-prototype component of EchoOOD (Sec.~\ref{subsec:ood_scoring}); EMA updates proceed irrespective of the gate.

\noindent\textbf{Normalization and local vs.\ global prototypes.}
For cosine-based operations used later, we employ numerically safe L2-normalized prototypes
$\bar{\mathbf{p}}^{\mathrm{g}}_k=\mathbf{p}^{\mathrm{g}}_k/\max\!\big(\|\mathbf{p}^{\mathrm{g}}_k\|_2,\epsilon\big)$
with a small $\epsilon>0$.
We distinguish these global EMA prototypes from the local scene-specific prototypes $\mathbf{p}^{\ell}_k$, which are constructed on-the-fly from confident predictions within a scene and are used by EchoOOD for local consistency scoring (Sec.~\ref{subsec:ood_scoring}).
The former provide stable class anchors across scenes, the latter capture scene-local coherence.

\subsection{Prototype-Guided Refinement}
\label{subsec:proto_refine}

\noindent\textbf{Prototype-Guided Semantic Imputation (PGSI).}
In unobserved regions, multiple geometrically plausible completions may exist, but only a subset is semantically valid. While conventional methods prioritize geometric consistency, they often overlook semantic coherence, which limits their generalization to OOD objects. We address this limitation by leveraging learned global class prototypes to guide feature imputation in unobserved regions (see Fig.~\ref{fig:pgocc}(b)).
We define the unobserved set as:
\begin{equation}
\mathcal{U}=\{\, i \mid \mathrm{occ}_i>\theta,\ i\notin\mathcal{V}\,\},
\end{equation}
where $\mathrm{occ}_i=\mathrm{sigmoid}(g(\mathbf{x}^{\text{coarse}}_i))$ is the occupancy probability from an auxiliary head $g(\cdot)$, $\theta\in(0,1)$ is a confidence threshold, and $\mathcal{V}$ is the set of visible voxels from the depth-guided proposal. For $i\in\mathcal{U}$ with coarse feature $\mathbf{x}^{\text{coarse}}_i$ (from $F^{3D}_{\text{coarse}}$), PGSI computes attention over mature global prototypes $\{\mathbf{p}^{\mathrm{g}}_k\}_{k=1}^{K_{\mathrm{cls}}}$:
\begin{equation}
a_{ik}=\frac{\exp(-\|\mathbf{x}^{\text{coarse}}_i-\mathbf{p}^{\mathrm{g}}_k\|^2/\tau_{\mathrm{att}})}{\sum_{k'=1}^{K_{\mathrm{cls}}}\exp(-\|\mathbf{x}^{\text{coarse}}_i-\mathbf{p}^{\mathrm{g}}_{k'}\|^2/\tau_{\mathrm{att}})}.
\end{equation}
We then apply a residual update with weight $\alpha_{\mathrm{pgsi}}\in(0,1]$:
\begin{equation}
\tilde{\mathbf{x}}^{\text{coarse}}_i
=\mathbf{x}^{\text{coarse}}_i
+\alpha_{\mathrm{pgsi}}\!\left(\sum_{k=1}^{K_{\mathrm{cls}}} a_{ik}\,\mathbf{p}^{\mathrm{g}}_k\right).
\end{equation}
This creates a self-reinforcing cycle: better completions yield more accurate prototypes, while improved prototypes enhance semantic reasoning in ambiguous regions.

\noindent\textbf{Prototype-Guided Tail Mining (PGTM).}
Enhancing modeling of tail classes is beneficial for overall performance.
We select tail-class candidates at the early semantic stage, where features remain coarse yet are already aligned with class prototypes (Fig.~\ref{fig:pgocc}(c)). Let $\mathcal{C}_{\text{tail}}$ be the set of rare classes. Using the PGSI-updated coarse features $\tilde{\mathbf{x}}^{\text{coarse}}_i$, we compute cosine similarity with global prototypes:
\begin{align}
s_{ik} &= \big\langle \bar{\tilde{\mathbf{x}}}^{\text{coarse}}_i,\ \bar{\mathbf{p}}^{\mathrm{g}}_k \big\rangle,\\
\bar{\tilde{\mathbf{x}}}^{\text{coarse}}_i &= \tilde{\mathbf{x}}^{\text{coarse}}_i \big/ \|\tilde{\mathbf{x}}^{\text{coarse}}_i\|_2.
\end{align}

Let $\max_k s_{ik}$ be the index of the largest value in $\{s_{ik}\}_{k=1}^{K_{\mathrm{cls}}}$, and
$\operatorname{second\_max}_k s_{ik}$ the second-largest value. We define the margin $\Delta_i=\max_k s_{ik}-\operatorname{second\_max}_k s_{ik}$ and select candidates via a two-stage filter followed by a Top-K operator:
\begin{equation}
\mathcal{I}_{\text{tail}}=\operatorname{TopK}\!\Big(\ \{\, i\mid \max_{k\in\mathcal{C}_{\text{tail}}} s_{ik}>\eta,\ \Delta_i>\delta \,\},\ K_{\mathrm{top}}\Big),
\end{equation}
with thresholds $\eta,\delta$ and selection size $K_{\mathrm{top}}\in\mathbb{N}$. For $i\in\mathcal{I}_{\text{tail}}$, we aggregate tail-class prototypes:
\begin{align}
w_{ik} &= \frac{\exp(s_{ik}/\tau_{\text{tail}})}{\sum_{k'\in\mathcal{C}_{\text{tail}}}\exp(s_{ik'}/\tau_{\text{tail}})},\\
\tilde{\mathbf{x}}^{\text{coarse}}_i
&\leftarrow \tilde{\mathbf{x}}^{\text{coarse}}_i
+ \psi\!\Big(\sum_{k\in\mathcal{C}_{\text{tail}}} w_{ik}\,\mathbf{p}^{\mathrm{g}}_k\Big),
\end{align}
where $\psi(\cdot)$ is a lightweight MLP and $\tau_{\text{tail}}>0$ is a temperature. A dedicated head $\phi(\cdot)$ predicts logits at $\mathcal{I}_{\text{tail}}$ and is supervised after the 3D backbone refinement by:
\begin{equation}
\mathcal{L}_{\text{tail}}=\frac{1}{|\mathcal{I}_{\text{tail}}|}\sum_{i\in\mathcal{I}_{\text{tail}}}\ell_{\text{CE}}\big(\phi(\mathbf{x}^{\text{refined}}_i),y_i\big),
\end{equation}
where $\mathbf{x}^{\text{refined}}_i$ denotes the feature of voxel $i$ in $F^{3D}_{\text{refined}}$ and $y_i$ is the ground-truth label. This coarse-to-refined strategy uses early semantic signals for candidate selection while grounding supervision on high-fidelity features, significantly improving rare-class performance.

\noindent\textbf{Prototype-Based Contrastive Learning (PBCL).}
To enforce global semantic consistency, we apply a prototype-based contrastive loss on $F^{3D}_{\text{refined}}$. Let $\mathcal{S}$ denote the set of non-empty voxels with ground-truth labels. For non-empty voxels $i\in\mathcal{S}$:
\begin{align}
s_{i,k} &= \big\langle \bar{\mathbf{x}}_i,\ \operatorname{sg}(\bar{\mathbf{p}}^{\mathrm{g}}_{k})\big\rangle, \quad k=1,\dots,K_{\mathrm{cls}},\\
\pi_{i,k} &= \frac{\exp(s_{i,k}/\tau_{\mathrm{cl}})}{\sum_{j=1}^{K_{\mathrm{cls}}}\exp(s_{i,j}/\tau_{\mathrm{cl}})},\\
\mathcal{L}_{\text{proto}} &= -\frac{1}{|\mathcal{S}|}\sum_{i\in\mathcal{S}} \log \pi_{i,y_i}.
\end{align}
where $\bar{\mathbf{x}}_i=\mathbf{x}_i/\|\mathbf{x}_i\|_2$, $\tau_{\mathrm{cl}}>0$ is a temperature, and $\operatorname{sg}(\cdot)$ stops gradients from flowing into prototypes. This loss is activated after prototype maturity is reached, promoting intra-class compactness and inter-class separation.

\begin{figure}[!t]
    \centering
    \includegraphics[width=\linewidth]{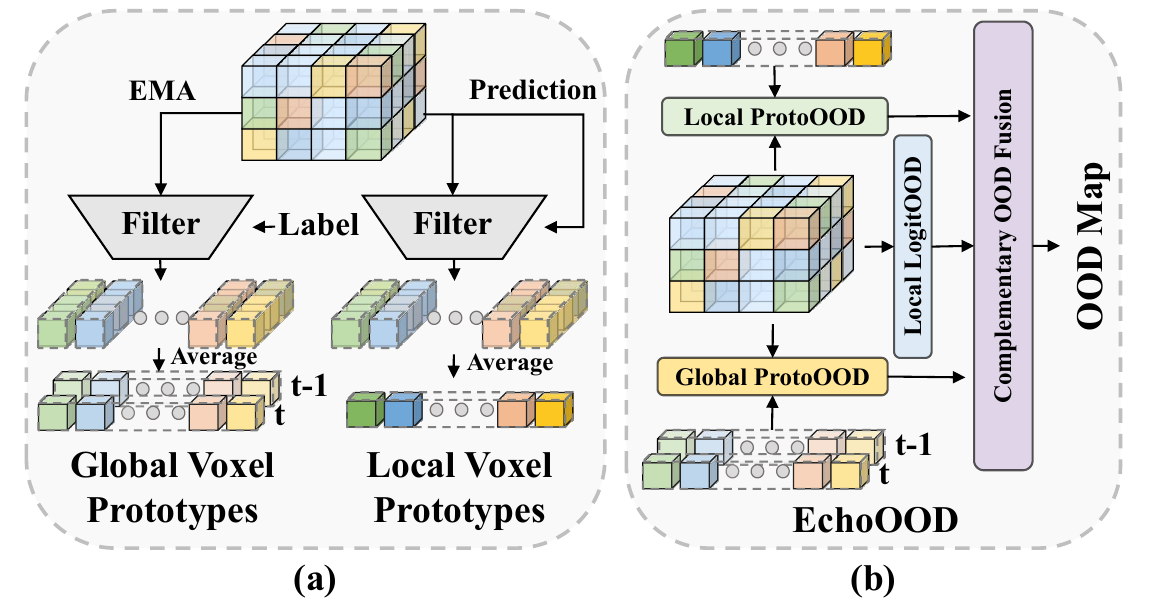}
    \vspace{-4ex}
    \caption{ \textbf{(a) Generation of Local and Global Voxel Prototypes.} \textbf{(b) EchoOOD.} It computes OOD maps via three cues: local logit alignment, and local/global prototype matching.}
    \label{fig:echoood} 
    \vspace{-3ex}
\end{figure}

\input{table/semantickitti_test}

\subsection{EchoOOD: Unsupervised Local \& Global Prototype Scoring}
\label{subsec:ood_scoring}

Detecting out-of-distribution regions is critical for safe and reliable 3D scene understanding, particularly in real-world deployments where novel objects may appear. Importantly, this challenge is exacerbated in long-tailed settings. 
Rare classes often exhibit low feature density and high intra-class variance, which can cause models to misattribute uncertain predictions to these underrepresented categories. 

Unlike conventional OOD detection methods that require auxiliary networks, we propose EchoOOD, a prototype-driven scoring mechanism within ProOOD that leverages the voxel prototypes (see Fig.~\ref{fig:echoood}(a)) already learned during occupancy prediction. ProOOD’s enhanced modeling of occluded regions and tail classes provides richer semantic priors, directly boosting EchoOOD’s detection performance. 
As illustrated in Fig.~\ref{fig:echoood}(b), EchoOOD computes three complementary OOD scores per voxel: Local prototype score (intra-class feature consistency), Local logit score (logit distribution coherence), and Global prototype score (alignment to canonical class prototypes). These scores are fused via a maximum-aggregation strategy to produce a final OOD map robust to long-tail biases.

Given a voxel-wise feature map $\mathbf{F}\in\mathbb{R}^{H\times W\times Z\times C}$ and its semantic prediction $\hat{\mathbf{Y}}\in\mathbb{R}^{H\times W\times Z}$ (here $\mathbf{F}=F^{3D}_{\text{refined}}$), EchoOOD computes the following scores for voxels with $\hat y_i>0$:

\noindent\textbf{Local logit score.}  
We measure logit coherence within each predicted class via the class-wise mean logit:
\begin{equation}
\begin{aligned}
\boldsymbol{\mu}_k &= \frac{1}{|\Omega_k|}\sum_{j\in\Omega_k} \mathbf{l}_j,\\
s_i^{\text{local-logit}} &= 1 - \frac{\langle \mathbf{l}_i, \boldsymbol{\mu}_{\hat{y}_i} \rangle}{\|\mathbf{l}_i\|_2\,\|\boldsymbol{\mu}_{\hat{y}_i}\|_2},
\end{aligned}
\end{equation}
where $\mathbf{l}_i \in \mathbb{R}^{K_{\mathrm{cls}}}$ denotes the logits for voxel $i$.

\noindent\textbf{Local prototype score.}  
For each predicted non-empty class $k$, we form a local scene-specific prototype by averaging features of confident voxels:
\begin{equation}
\begin{aligned}
\mathbf{p}^{\ell}_k &= \frac{1}{|\Omega_k|}\sum_{j\in\Omega_k} \mathbf{x}_j,\\
\Omega_k &= \{\, j \mid \hat{y}_j = k,\ \mathrm{gap}_j > \tau_{\mathrm{conf}} \,\},
\end{aligned}
\end{equation}
where $\hat{y}_j$ is the predicted class label of voxel $j$, $\mathrm{gap}_j = p_j^{(1)} - p_j^{(2)}$ is the softmax top-2 probability gap of $\mathbf{l}_j$, and $\tau_{\mathrm{conf}}\in(0,1)$ is a confidence threshold. The local prototype score is:
\begin{equation}
s_i^{\text{local-proto}} 
= 1 - \frac{\langle \mathbf{x}_i, \mathbf{p}^{\ell}_{\hat{y}_i} \rangle}{\|\mathbf{x}_i\|_2 \cdot \|\mathbf{p}^{\ell}_{\hat{y}_i}\|_2}.
\end{equation}

\noindent\textbf{Global prototype score.}  
We evaluate alignment between each voxel and its corresponding global EMA prototype:
\begin{equation}
s_i^{\text{global-proto}} 
= 1 - \frac{\langle \mathbf{x}_i, \mathbf{p}^{\mathrm{g}}_{\hat{y}_i} \rangle}{\|\mathbf{x}_i\|_2 \cdot \|\mathbf{p}^{\mathrm{g}}_{\hat{y}_i}\|_2}.
\end{equation}

\noindent\textbf{Fusion and empty voxels.}  
Each component score is min–max normalized to $[0,1]$ over the scene before fusion. The final score uses maximum aggregation:
\begin{equation}
s_i^{\text{fused}} 
= \max\!\left( s_i^{\text{local-logit}},\ s_i^{\text{global-proto}},\ s_i^{\text{local-proto}} \right),
\end{equation}
where voxels predicted as empty (class $0$) are assigned $s_i^{\text{fused}} = \min_{j} s_j^{\text{fused}}$, \textit{i.e.}, the minimum fused score across the entire scene.

EchoOOD requires no additional parameters or training. It directly exploits the semantic prototypes maintained for occupancy prediction, providing an efficient plug-and-play solution for OOD detection in 3D occupancy prediction.

%% file: table/semantickitti_test.tex
\begin{table*}[!htb]
    \centering
    \resizebox{\textwidth}{!}{%
        \fontsize{7.6pt}{11pt}\selectfont
        \setlength{\tabcolsep}{2.5pt}
        \begin{tabular}{@{}l|c|*{19}{c}|c|c@{}}
            \toprule
            Method 
            & IoU
            & \rotatebox{90}{{\textcolor{red}{\rule{0.5em}{0.5em}}}\hspace{0.3em}road \scalebox{0.7}{(15.30\%)}} 
            & \rotatebox{90}{{\textcolor{pink}{\rule{0.5em}{0.5em}}}\hspace{0.3em}sidewalk \scalebox{0.7} {(11.13\%)}} 
            & \rotatebox{90}{{\textcolor{orange}{\rule{0.5em}{0.5em}}}\hspace{0.3em}parking \scalebox{0.7}{(1.12\%)}} 
            & \rotatebox{90}{{\textcolor{yellow}{\rule{0.5em}{0.5em}}}\hspace{0.3em}other-grnd. \scalebox{0.7}{(0.56\%)}} 
            & \rotatebox{90}{{\textcolor{green}{\rule{0.5em}{0.5em}}}\hspace{0.3em}building \scalebox{0.7}{(14.1\%)}} 
            & \rotatebox{90}{{\textcolor{blue}{\rule{0.5em}{0.5em}}}\hspace{0.3em}car \scalebox{0.7}{(3.92\%)}} 
            & \rotatebox{90}{{\textcolor{purple}{\rule{0.5em}{0.5em}}}\hspace{0.3em}truck \scalebox{0.7}{(0.16\%)}} 
            & \rotatebox{90}{{\textcolor{brown}{\rule{0.5em}{0.5em}}}\hspace{0.3em}bicycle \scalebox{0.7}{(0.03\%)}} 
            & \rotatebox{90}{{\textcolor{cyan}{\rule{0.5em}{0.5em}}}\hspace{0.3em}motorcycle \scalebox{0.7}{(0.03\%)}} 
            & \rotatebox{90}{{\textcolor{magenta}{\rule{0.5em}{0.5em}}}\hspace{0.3em}other-veh. \scalebox{0.7}{(0.20\%)}} 
            & \rotatebox{90}{{\textcolor{lime}{\rule{0.5em}{0.5em}}}\hspace{0.3em}vegetation \scalebox{0.7}{(39.3\%)}} 
            & \rotatebox{90}{{\textcolor{teal}{\rule{0.5em}{0.5em}}}\hspace{0.3em}trunk \scalebox{0.7}{(0.51\%)}} 
            & \rotatebox{90}{{\textcolor{violet}{\rule{0.5em}{0.5em}}}\hspace{0.3em}terrain \scalebox{0.7}{(9.17\%)}} 
            & \rotatebox{90}{{\textcolor{gray}{\rule{0.5em}{0.5em}}}\hspace{0.3em}person \scalebox{0.7}{(0.07\%)}} 
            & \rotatebox{90}{{\textcolor{black}{\rule{0.5em}{0.5em}}}\hspace{0.3em}bicyclist \scalebox{0.7}{(0.07\%)}} 
            & \rotatebox{90}{{\textcolor{lightgray}{\rule{0.5em}{0.5em}}}\hspace{0.3em}motorcyclist \scalebox{0.7}{(0.05\%)}} 
            & \rotatebox{90}{{\textcolor{olive}{\rule{0.5em}{0.5em}}}\hspace{0.3em}fence \scalebox{0.7}{(3.90\%)}}
            & \rotatebox{90}{{\textcolor{navy}{\rule{0.5em}{0.5em}}}\hspace{0.3em}pole \scalebox{0.7}{(0.29\%)}} 
            & \rotatebox{90}{{\textcolor{maroon}{\rule{0.5em}{0.5em}}}\hspace{0.3em}traf.-sign \scalebox{0.7}{(0.08\%)}} 
            & mIoU & \shortstack{\emph{Tail} \\ mIoU}\\
            \midrule 
            \multicolumn{22}{c}{\textbf{\emph{SemanticKITTI}}} \\
            \midrule
            \hline
            MonoScene~\cite{cao2022monoscene}  
            &34.16 &54.70 &27.10 &24.80 &5.70 &14.40 &18.80 &3.30 &0.50 &0.70 &4.40 &14.90 &2.40 &19.50 &1.00 &1.40 &0.40 &11.10 &3.30 &2.10 &11.08 &2.29\\
            
            TPVFormer~\cite{huang2023tri}  
            &34.25  &55.10 &27.20 &27.40 &6.50 &14.80 &19.20 &3.70 &1.00 &0.50 &2.30 &13.90 &2.60 &20.40 &1.10 &2.40 &0.30 &11.00 &2.90 &1.50 &11.26 &2.25\\

            OccFormer~\cite{zhang2023occformer}  
            &34.53 &55.90 &30.30 &31.50 &6.50 &15.70 &21.60 &1.20 &1.50 &1.70 &3.20 &16.80 &3.90 &21.30 &2.20 &1.10 &0.20 &11.90 &3.80 &3.70 &12.32 &2.64\\

            VoxFormer~\cite{li2023voxformer}  
            & 42.95 & 53.90 &25.30 &21.10 &5.60 &19.80 &20.80 &3.50 &1.00 &0.70 &3.70 &22.40 &7.50 &21.30 &1.40 &2.60 &0.20 &11.10 &5.10 &4.90 & 12.20 &3.29\\

            HASSC~\cite{wang2024not}  
            & 43.40 & 54.60 &27.70 &23.80 &6.20 &21.10 &22.80 &4.70 &1.60 &1.00 &3.90 &23.80 &8.50 &23.30 &1.60 &\textbf{4.00} &0.30 &13.10 &5.80 &5.50 & 13.34 &3.92\\
            Symphonies~\cite{jiang2024symphonize}
            & 42.19 & 58.40 & 29.30 & 26.90 & 11.70 & 24.70 & 23.60 & 3.20 & 3.60 & 2.60 & 5.60 & 24.20 & 10.00 & 23.10 & 3.20 & 1.90 & \textbf{2.00} & 16.10 & 7.70 & 8.00 & 15.04 &5.41\\    
            
            SHTOcc~\cite{yu2025shtocc}
            & - & 57.80 &28.90 &26.30 &12.00 &23.40 &24.60 &3.60 &3.50 &1.80 &4.60 &24.90 &11.00 &22.50 &\textbf{3.70} &3.80 &0.10 &17.10 &8.70 &9.40 & 15.07 &5.65\\    

            CGFormer~\cite{yu2024context} 
            & 44.41 & 64.30 & 34.20 & 34.10 & \textbf{12.10} & 25.80 & 26.10 & 4.30 & 3.70 & 1.30 & 2.70 & 24.50 & 11.20 & 29.30 & 1.70 & 3.60 & 0.40 & 18.70 & 8.70 & 9.30 & 16.63 &5.36\\            

            \midrule   
            SGN~\cite{mei2024camera}
            & 41.88 & \textbf{57.80} & \textbf{29.20} & 27.70 & 5.20 & \textbf{23.90} & \textbf{24.90} & 2.70 & 0.40 & 0.30 & \textbf{4.00} & 24.20 & 10.00 & \textbf{25.80} & 1.10 & \textbf{2.50} & 0.30 & 14.20 & 7.40 & 4.40 & 14.01 &3.48\\
            
            \rowcolor{mygray}ProOOD (+SGN)~\cite{mei2024camera}
            & \textbf{43.14} & 57.10 & 28.00 & \textbf{28.00} & \textbf{7.10} & 23.80 & 24.70 & \textbf{4.20} & \textbf{1.30} & \textbf{0.90} & 2.40 & \textbf{25.20} & \textbf{11.40} & 24.80 & \textbf{2.10} & 2.00 & \textbf{0.30} & \textbf{16.20} & \textbf{8.10} & \textbf{8.00} & \textbf{14.51} &\textbf{4.35}\\

            \midrule   
            VoxDet*~\cite{yu2024context} 
            & 46.69 & 65.80 &35.50 &33.90 &11.40 &27.10 &27.50 &\textbf{5.20} &\textbf{5.10} &\textbf{5.60} &4.50 &27.70 &11.90 &31.20 &2.30 &2.70 &\textbf{0.50} &21.10 &9.50 &9.20 & 17.77  &6.17\\
            \rowcolor{mygray}ProOOD (+VoxDet)
            & \textbf{46.75} & \textbf{66.20} & \textbf{36.40}& \textbf{34.90} & \textbf{11.40} & \textbf{27.70} & \textbf{27.60} & 4.30 & 4.10 &  5.00& \textbf{7.20} & \textbf{28.40} & \textbf{12.90}& \textbf{32.20} &\textbf{2.50} & \textbf{2.80} & 0.40 & \textbf{21.30} & \textbf{9.60} & \textbf{9.50}& \textbf{18.12}&\textbf{6.34}

            \\
            \midrule
            \multicolumn{22}{c}{\textbf{\emph{VAA-KITTI}}}\\
            \midrule
            \hline
            SGN~\cite{mei2024camera}&39.07  &46.62 &26.54 &15.88 &0.12 &18.34 &\textbf{20.51} &0.62 &\textbf{0.44} &0.35 &1.23 &24.76 &6.40 &21.78&0.48 &\textbf{1.29} &0.00 &12.25 &8.70 &4.57&11.10&2.20
            \\                
            \rowcolor{mygray}ProOOD (+SGN)&\textbf{40.26}  &\textbf{49.34} &\textbf{27.14} &\textbf{16.47} &\textbf{0.15} &\textbf{20.07} &19.94 &\textbf{3.32} &0.43 &\textbf{0.94} &\textbf{1.31} &\textbf{26.08} &\textbf{6.50} &\textbf{23.10}&\textbf{1.57} &1.18 &0.00 &\textbf{12.89} &\textbf{9.06} &\textbf{5.87}&\textbf{11.86}&  \textbf{2.76}\\             
            \midrule         
            VoxDet~\cite{li2025voxdet}&\textbf{41.60}  &55.93 & 27.12&21.33 &0.02 &\textbf{22.81} &20.38 &5.77& \textbf{3.27}&1.08 & \textbf{4.14}&\textbf{27.36} & 6.03& \textbf{23.13}& 1.09&1.70 & 0.00& 15.73& 8.83&4.68 &13.18 &3.33
            \\                
            \rowcolor{mygray}ProOOD (+VoxDet)& 40.88 &\textbf{56.41} & \textbf{28.20}&\textbf{23.77} & \textbf{0.19}& 21.73&\textbf{21.32} &\textbf{7.00} &2.89 &\textbf{3.78} & 3.57& 27.07&\textbf{6.41} &21.21 & \textbf{2.62}& \textbf{1.98}& 0.00& \textbf{15.96}& \textbf{9.06}&\textbf{6.05} & \textbf{13.63} &\textbf{3.96}\\         
            \bottomrule
        \end{tabular}%
    }
    \caption{Quantitative results on the SemanticKITTI test set~\cite{behley2019semantickitti} and VAA-KITTI~\cite{zhang2025occood} (Single-Frame Evaluation for Fair Comparison). * denotes results reproduced using the official code implementation.}
    \vspace{-3ex}
    \label{tab:kitti}
\end{table*}

%% file: sec/4_experiments.tex
\section{Experiments}
\label{sec:experiments}
\input{table/VAA-KITTI_OOD}
\input{table/SSCBench-KITTI360_test}

\subsection{Experiment Setup}
\noindent\textbf{Datasets.} 
We evaluate ProOOD on five datasets: SemanticKITTI~\cite{behley2019semantickitti} and SSCBench-KITTI-360~\cite{li2024sscbench} for 3D occupancy prediction; VAA-KITTI~\cite{zhang2025occood}, VAA-KITTI-360~\cite{zhang2025occood}, and VAA-STU~\cite{zhang2025occood} for out-of-distribution detection following OccOoD~\cite{zhang2025occood}. 
SemanticKITTI provides RGB images ($1226{\times}370$) and occupancy labels with $20$ semantic classes ($19$ objects + \emph{free} space), split into $10$ train, $1$ val, and $11$ test sequences. SSCBench-KITTI-360 offers $1408{\times}376$ RGB images and labels with $19$ classes ($18$ objects + free space), using $7$ train, $1$ val, and $1$ test sequences. 
VAA-KITTI~\cite{zhang2025occood} and VAA-KITTI-360~\cite{zhang2025occood} are derived from SemanticKITTI and SSCBench-KITTI-360, respectively, each containing $500$ test images with synthetic anomalies. 
VAA-STU~\cite{zhang2025occood} is constructed from two high-anomaly sequences of STU~\cite{nekrasov2025stu}, a real-world 3D point cloud dataset containing genuine anomalies, yielding $578$ test images. The point cloud ground truths are voxelized to facilitate evaluation in the 3D occupancy space. 
For more details, please refer to the supplement.

\vspace{3pt}\noindent\textbf{Evaluation Metrics.}
For 3D occupancy prediction, we report IoU and mIoU following standard protocols, supplemented by \emph{tail} mIoU to assess performance on underrepresented classes.
In SemanticKITTI, tail classes (\emph{e.g.}, \emph{truck}: $0.16\%$, \emph{traffic-sign}: $0.08\%$) collectively account for $2\%$ of total voxels. In SSCBench-KITTI-360, classes include \emph{bicycle} ($0.01\%$) and \emph{other-object} ($0.28\%$), comprising $0.8\%$ voxels of the dataset. Please see the supplement for the full class distributions. For OOD detection, we adopt the metrics from OccOoD~\cite{zhang2025occood}: $AuPRC_r$ and $AuROC$, for performance evaluation. $AuPRC_r$ incorporates spatial tolerance via ground-truth dilation at $0.8m$, $1.0m$, and $1.2m$; $AuROC$ captures global separation quality.

\input{table/ablation_occ}
\input{table/ablation_ood}

\vspace{3pt}\noindent\textbf{Implementation Details.} 
We integrate ProOOD into several state-of-the-art occupancy methods, \textit{e.g.}, SGN~\cite{mei2024camera} and VoxDet~\cite{li2025voxdet}, following their original training configurations. All models are reimplemented using $4$ NVIDIA RTX 3090 GPUs. 
To accommodate GPU memory constraints, we adjust the batch size per GPU, such as reducing from the original setting in VoxDet~\cite{li2025voxdet}, which used $2$ A100 GPUs.
Our method only introduces $0.28M$ additional parameters, preserving efficiency while boosting the OOD detection capability. More details of implementation and hyperparameters are provided in the supplementary material.

\subsection{Experimental Results}
\noindent\textbf{3D Occupancy Prediction.} 
As shown in Tab.~\ref{tab:kitti} and Tab.~\ref{table:kitti-360}, we present the experimental results of 3D occupancy prediction. Our method consistently outperforms the baseline model while significantly improving performance on tail classes. For instance, on the SemanticKITTI dataset compared to SGN~\cite{mei2024camera},  ProOOD achieves a substantial $24.8\%$ improvement in mIoU for tail categories. Moreover, on OccOoD benchmarks, OOD objects severely degrade model performance, particularly for tail classes. ProOOD handles such cases more effectively and exhibits stronger robustness, thanks to its prototype-guided design that enhances semantic plausibility in occluded regions and improves modeling of tail classes. On VAA-KITTI, for example, it improves tail mIoU by ${+}0.56\%$ over the baseline.

\vspace{3pt}\noindent\textbf{Out-of-Distribution Detection.}
OccOoD, inherently built upon 3D occupancy prediction, requires balancing in-distribution confidence with overall accuracy.
Our proposed method achieves new state-of-the-art performance not only in 3D occupancy prediction, but also in out-of-distribution detection, as shown in Tab.~\ref{tab:vaa-kitti}, with consistent gains in $AuPRC_{r}$ across VAA-KITTI, VAA-KITTI-360, and the real-world anomaly dataset: VAA-STU. These gains arise from improved long-tail modeling during occupancy prediction, which mitigates \textit{strong classification} and enhances the sensitivity to OOD voxels, alongside the incorporation of both local and global prototypes in EchoOOD, which enable more precise identification of OOD voxels.

\vspace{3pt}\noindent\textbf{Visualization.} 
As shown in Fig.~\ref{fig:comparsion}, we compare the occupancy and OOD visualization results of SGN~\cite{mei2024camera} and the proposed method. SGN fails to accurately predict the occupancy state of the OOD object ``\emph{clothing}'', retaining only a few scattered voxels and assigning it a low anomaly score. In contrast, benefiting from the proposed PGSI module, ProOOD precisely reconstructs the occupancy state, while the PGTM further strengthens sensitivity to OOD objects, resulting in a significantly higher anomaly score for it. More qualitative results are in the supplement.

\subsection{Ablation Studies}
In this section, we evaluate the effectiveness of each component and different anomaly score methods. More ablation studies are shown in the supplement.

\vspace{3pt}\noindent\textbf{Components.}
To dissect the contribution of each component, we conduct an ablation study as presented in Tab.~\ref{tab:ablation_components}. By introducing the Prototype-Based Contrastive Loss (PBCL), \emph{tail} mIoU increases by $9.09\%$, validating its role in enhancing intra-class compactness and inter-class separation. Incorporating PGSI improves IoU, demonstrating its effectiveness in recovering occluded regions through voxel prototypes. Furthermore, the PGTM increases mIoU by $1.42\%$ and \emph{tail} mIoU by $6.77\%$, highlighting its ability to effectively mine underrepresented tail-class voxels.

\vspace{3pt}\noindent\textbf{Anomaly Score Methods.} 
Tab.~\ref{tab:ood_vaa-kitti} demonstrates the effectiveness of our proposed EchoOOD. Compared to the other scoring methods, EchoOOD improves $AuPRC_r$ by more than $12\%$ across all scales, while maintaining comparable $AuROC$. This validates that explicitly modeling long-tail categories robustly enhances OOD voxel detection, and further confirms the consistent efficacy of leveraging both local and global voxel prototypes for OOD detection.

%% file: table/VAA-KITTI_OOD.tex
\begin{table*}[t]
        \centering
        \small
        \resizebox{\textwidth}{!}{
        \begin{tabular}{@{}l|c|ccc|c|ccc|c|ccc|c@{}}
        \toprule
        & &\multicolumn{4}{c|}{VAA-KITTI}&\multicolumn{4}{c|}{VAA-KITTI-360}&\multicolumn{4}{c}{VAA-STU}\\
        \multirow{2}{*}{Methods} &\multirow{2}{*}{\makecell[c]{Anomaly Score \\ Methods}}& \multicolumn{3}{c|}{AuPRC$_r$$\uparrow$} & \multirow{2}{*}{AuROC$\uparrow$}&\multicolumn{3}{c|}{AuPRC$_r$$\uparrow$} & \multirow{2}{*}{AuROC$\uparrow$}& \multicolumn{3}{c|}{AuPRC$_r$$\uparrow$} & \multirow{2}{*}{AuROC$\uparrow$}\\
         & & $0.8m$ & $1.0m$ & $1.2m$ &  & $0.8m$ & $1.0m$ & $1.2m$ &  & $0.8m$ & $1.0m$ & $1.2m$ & \\
        \midrule
        VoxDet~\cite{li2025voxdet}& \multirow{3}{*}{ASS~\cite{zhang2025occood}}&  4.57&8.36 & 14.02 & 59.52  &11.79  & 25.08 & 48.49 &\textbf{50.76}  & 1.45 & 3.05 & 5.89 &59.52\\ 
        SGN~\cite{mei2024camera} &&8.22 &14.45 & 22.93& 60.51 & 13.27 & 28.57 & 52.99 &47.27  & 2.91&5.74  & 11.32 &64.51\\
        OccOoD~\cite{zhang2025occood}  &&10.80 & 16.05 & 23.42 & 61.96 &14.26  & 29.72 & 54.35 & 47.84& 3.05 & 6.15 & 11.39 &\textbf{65.37}\\   

        \midrule
        Ours (+VoxDet)  & \multirow{2}{*}{EchoOOD}&21.95 & 36.97 &  56.49 & 60.99 & 12.75 & 26.81 &  50.08& 47.24 & 2.74 & 5.09 & 8.82 &55.11\\ 
        Ours (+SGN)& &\textbf{27.86}& \textbf{43.55}&  \textbf{62.65}   &  \textbf{64.31}& \textbf{14.56}& \textbf{30.83} & \textbf{57.50} & 46.91 & \textbf{7.98} & \textbf{14.54} & \textbf{24.21} &62.11\\ 

        \bottomrule
        \end{tabular}}
        \caption{Quantitative OOD results on VAA-KITTI, VAA-KITTI-360, and VAA-STU datasets~\cite{zhang2025occood}.}
        \vskip -2ex
        \label{tab:vaa-kitti}
\end{table*}

%% file: table/SSCBench-KITTI360_test.tex
\begin{table*}[!t]
    \centering


    \resizebox{\textwidth}{!}{
        \fontsize{7.8pt}{10pt}\selectfont
        \setlength{\tabcolsep}{2.5pt} 
        \begin{tabular}{@{}l|c|*{18}{c}|c|c@{}}
            \toprule
            Method 
            &IoU
            & \rotatebox{90}{{\textcolor{blue}{\rule{0.5em}{0.5em}}}\hspace{0.3em}car \scalebox{0.7}{(2.85\%)}} 
            & \rotatebox{90}{{\textcolor{brown}{\rule{0.5em}{0.5em}}}\hspace{0.3em}bicycle \scalebox{0.7}{(0.01\%)}} 
            & \rotatebox{90}{{\textcolor{cyan}{\rule{0.5em}{0.5em}}}\hspace{0.3em}motorcycle \scalebox{0.7}{(0.01\%)}} 
            & \rotatebox{90}{{\textcolor{purple}{\rule{0.5em}{0.5em}}}\hspace{0.3em}truck \scalebox{0.7}{(0.16\%)}} 
            & \rotatebox{90}{{\textcolor{magenta}{\rule{0.5em}{0.5em}}}\hspace{0.3em}other-veh. \scalebox{0.7}{(5.75\%)}} 
            & \rotatebox{90}{{\textcolor{gray}{\rule{0.5em}{0.5em}}}\hspace{0.3em}person \scalebox{0.7}{(0.02\%)}} 
            & \rotatebox{90}{{\textcolor{red}{\rule{0.5em}{0.5em}}}\hspace{0.3em}road \scalebox{0.7}{(14.98\%)}} 
            & \rotatebox{90}{{\textcolor{orange}{\rule{0.5em}{0.5em}}}\hspace{0.3em}parking \scalebox{0.7}{(2.31\%)}}           
            & \rotatebox{90}{{\textcolor{pink}{\rule{0.5em}{0.5em}}}\hspace{0.3em}sidewalk \scalebox{0.7}{(6.43\%)}} 
            & \rotatebox{90}{{\textcolor{yellow}{\rule{0.5em}{0.5em}}}\hspace{0.3em}other-grnd. \scalebox{0.7}{(2.05\%)}} 
            & \rotatebox{90}{{\textcolor{green}{\rule{0.5em}{0.5em}}}\hspace{0.3em}building \scalebox{0.7}{(15.67\%)}} 
            & \rotatebox{90}{{\textcolor{olive}{\rule{0.5em}{0.5em}}}\hspace{0.3em}fence \scalebox{0.7}{(0.96\%)}} 
            & \rotatebox{90}{{\textcolor{lime}{\rule{0.5em}{0.5em}}}\hspace{0.3em}vegetation \scalebox{0.7}{(41.99\%)}} 
            & \rotatebox{90}{{\textcolor{violet}{\rule{0.5em}{0.5em}}}\hspace{0.3em}terrain \scalebox{0.7}{(7.10\%)}} 
            & \rotatebox{90}{{\textcolor{navy}{\rule{0.5em}{0.5em}}}\hspace{0.3em}pole \scalebox{0.7}{(0.22\%)}} 
            & \rotatebox{90}{{\textcolor{maroon}{\rule{0.5em}{0.5em}}}\hspace{0.3em}traf.sign \scalebox{0.7}{(0.06\%)}} 
            & \rotatebox{90}{{\textcolor{blue}{\rule{0.5em}{0.5em}}}\hspace{0.3em}other-struct. \scalebox{0.7}{(4.33\%)}}             
            & \rotatebox{90}{{\textcolor{teal}{\rule{0.5em}{0.5em}}}\hspace{0.3em}other-obj. \scalebox{0.7}{(0.28\%)}} 

            & mIoU &\shortstack{\emph{Tail} \\ mIoU}\\
            \midrule
            \multicolumn{21}{c}{\textbf{\emph{SSCBench-KITTI-360}}}\\
            \midrule
            \hline
            MonoScene~\cite{cao2022monoscene} &37.87  &19.34 &0.43 &0.58 &8.02 &2.03 &0.86 &48.35 &11.38 &28.13 &3.32 &32.89 &3.53 &26.15 &16.75 &6.92 &5.67 &4.20 &3.09 &12.31 &3.61\\
            
            TPVFormer~\cite{huang2023tri} & 40.22  & 21.56 & 1.09 & 1.37 & 8.06 & 2.57 & 2.38 & 52.99 & 11.99 & 31.07 & 3.78 & 34.83 & 4.80 & 30.08 & 17.52 & 7.46 & 5.86 & 5.48 & 2.70 & 13.64&4.13\\
            VoxFormer~\cite{li2023voxformer} & 38.76  & 17.84 & 1.16 & 0.89 & 4.56 & 2.06 & 1.63 & 47.01 & 9.67 & 27.21 & 2.89 & 31.18 & 4.97 & 28.99 & 14.69 & 6.51 & 6.92 & 3.79 & 2.43  & 11.91&3.44\\     
  
            OccFormer~\cite{zhang2023occformer} & 40.27  & 22.58 & 0.66 & 0.26 & 9.89 & 3.82 & 2.77 & 54.30 & 13.44 & 31.53 & 3.55 & 36.42 & 4.80 & 31.00 & 19.51 & 7.77 & 8.51 & 6.95 & 4.60& 13.81&4.92\\
            Symphonies~\cite{jiang2024symphonize} &44.12  &\textbf{30.02} &1.85 &5.90 &\textbf{25.07} &\textbf{12.06} &\textbf{8.20} &54.94 &13.83 &32.76 &\textbf{6.93} &35.11 &8.58 &38.33 &11.52 &14.01 &9.57 &\textbf{14.44} &\textbf{11.28}  &18.58&10.84\\  
            \midrule
            SGN~\cite{mei2024camera} &46.22 &\textbf{28.20} &2.09 &3.02 &11.95 &3.68 &4.20 &59.49 &\textbf{14.50} &36.53 &4.24 &39.79 &\textbf{7.14} &36.61 &\textbf{23.10} &\textbf{14.86} &\textbf{16.14} &8.24 &\textbf{4.95} &17.71&8.17\\ 
            \rowcolor{mygray}ProOOD (+SGN)&\textbf{46.62}& 27.23& \textbf{2.75}&\textbf{4.23} &\textbf{12.06} &\textbf{5.76} &\textbf{4.83} & \textbf{60.07}&14.02 &  \textbf{36.57}&\textbf{5.03} &\textbf{40.92} &7.13 &\textbf{36.96} &22.84 &14.15 &15.47 & \textbf{8.82}&4.93 &\textbf{17.99}&\textbf{8.35}\\
            \midrule
            VoxDet*~\cite{li2025voxdet} &48.22 &\textbf{28.94} &\textbf{3.88} & 5.23&16.20 &\textbf{6.58} & \textbf{6.15}&62.02 & \textbf{16.27}&39.69 &5.34 &43.36 & 9.10& \textbf{38.73}& \textbf{22.79}&17.22 & \textbf{19.74}& 10.27&7.99 &19.97&10.92\\     
            \rowcolor{mygray}ProOOD (+VoxDet)&\textbf{48.23} & 28.89&3.77 & \textbf{7.23}&\textbf{18.83} & 6.49& 5.83& \textbf{62.31}& 16.23&\textbf{40.08} & \textbf{5.36}&\textbf{43.75} & \textbf{9.29}& 38.66&22.51 &\textbf{17.28} & 19.42& \textbf{10.67}& \textbf{8.25}&\textbf{20.27}&\textbf{11.52}\\            
            \midrule
            \multicolumn{21}{c}{\textbf{\emph{VAA-KITTI-360}}}\\
            \midrule
            \hline
            SGN~\cite{mei2024camera} &\textbf{39.68} &20.68 &0.05 &0.00 &0.00 &\textbf{2.92} &0.00 &47.59 &14.66 &24.71 &2.56 &32.83 &\textbf{3.52} &\textbf{28.68}&11.63 &7.44 &8.67 &4.29&\textbf{5.52} &11.99&3.10
            \\            
            \rowcolor{mygray}ProOOD (+SGN)& 39.38& \textbf{22.17}& \textbf{0.05}&0.00 &0.00 &0.05 &\textbf{2.10} &\textbf{48.19} &\textbf{16.50} &\textbf{25.48} & \textbf{3.62}&\textbf{33.75} &2.89 &28.08 &\textbf{12.86} & \textbf{8.78} & \textbf{10.29}&\textbf{5.57} &4.71 &\textbf{12.52}&\textbf{3.71}
            \\       
            \midrule
            VoxDet~\cite{li2025voxdet} &42.22 & 21.38& \textbf{0.99}& 0.00& 0.00&\textbf{0.37} & 2.61&51.41 &12.19 & \textbf{28.84}& 2.68& 36.19&3.50 & \textbf{29.86}& \textbf{12.54}& 9.30&15.83 & 5.46&8.35 &13.42&5.29
            \\            
            \rowcolor{mygray}ProOOD (+VoxDet)&\textbf{42.54}& \textbf{23.11}& 0.17& 0.00& 0.00&0.07 &\textbf{3.60} & \textbf{51.49}& \textbf{14.49}& 28.51& \textbf{3.60}&\textbf{37.31} & \textbf{4.01}&29.79 & 12.47& \textbf{10.64}&\textbf{19.01} & \textbf{7.17}&\textbf{9.83} &\textbf{14.18} &\textbf{6.18}
            \\  
            \bottomrule
        \end{tabular}
    }
    \caption{Quantitative results on the SSCBench-KITTI-360 test set~\cite{li2024sscbench} and VAA-KITTI-360~\cite{zhang2025occood} (Single-Frame Evaluation for Fair Comparison). * denotes results reproduced using the official code implementation.}
    \label{table:kitti-360}
    \vspace{-2ex}
\end{table*}

%% file: table/ablation_occ.tex
\begin{table}[]
    \centering
    \small
    \resizebox{0.48\textwidth}{!}{
    \begin{tabular}{@{}cccc|c|c|c@{}}
    \toprule
    Baseline & PBCL & PGSI & PGTM & IoU& mIoU& \emph{Tail} mIoU\\
    \midrule
    \ding{51} & & & & 43.60& 14.55& 4.07\\
    \ding{51}& \ding{51}& & & 43.69&14.63& 4.44 \textcolor{codegreen}{(+9.09\%)} \\
    \ding{51}&\ding{51} &\ding{51} & & 44.09&14.76&4.58 \textcolor{codegreen}{(+3.15\%)}\\
    \ding{51}&\ding{51} & \ding{51}& \ding{51}& 44.02&14.97& 4.89 \textcolor{codegreen}{(+6.77\%)}\\
     \bottomrule
    \end{tabular}}
    \vspace{-2ex}
    \caption{Ablation study of each component on the SemanticKITTI~\cite{behley2019semantickitti} validation set for 3D Occupancy prediction.}
    \vspace{-2ex}
    \label{tab:ablation_components}
\end{table}

%% file: table/ablation_ood.tex
\begin{table}[t]
    \centering
    \small
    \begin{tabular}{@{}c|ccc|c@{}}
        \toprule
        \multirow{3}{*}{\makecell{Anomaly \\ Score Methods}} & \multicolumn{4}{c}{VAA-KITTI}\\
        
        & \multicolumn{3}{c|}{AuPRC$_r$$\uparrow$} & \multirow{2}{*}{AuROC$\uparrow$} \\
        
         & $0.8m$ & $1.0m$ & $1.2m$ &\\
        \midrule

        Postpro & 5.81  & 10.02 & 16.22 & 63.85 \\
        Energy & 6.80& 12.11 & 20.08 & \textbf{66.69}\\
        Entropy&7.38  & 12.91 & 21.05 &66.24\\
        ASS~\cite{zhang2025occood} & 14.47 & 20.96 & 29.65 & 64.20\\
        \midrule
        EchoOOD (Ours) &\textbf{27.86}& \textbf{43.55}&  \textbf{62.65}   &  64.31 \\
        \bottomrule
    \end{tabular} 
        \vspace{-2ex}
        \caption{Ablation studies of anomaly score methods on the VAA-KITTI dataset~\cite{zhang2025occood} for OccOoD evaluation.}
        \label{tab:ood_vaa-kitti}
    \vspace{-3ex}
\end{table}

%% file: sec/5_conclusion.tex
\section{Conclusion}
\label{sec:conclusion}
In this paper, we introduce ProOOD, a plug-and-play method that enhances representations from a voxel prototype-guided perspective to simultaneously improve both 3D occupancy prediction and out-of-distribution detection.
ProOOD improves reconstruction in occluded regions through learned voxel prototypes and explicitly identifies latent long tail voxels, enabling us to investigate the impact of long tail categories on out-of-distribution 3D occupancy prediction.
We further introduce EchoOOD, a lightweight prototype-based scoring mechanism for OOD detection, which effectively discriminates novel OOD voxels. 
Extensive experiments demonstrate that ProOOD achieves state-of-the-art performance on both 3D occupancy prediction and OOD detection across multiple benchmarks.

%% file: sec/ack.tex
\section*{Acknowledgment}

This work was supported in part by the National Natural Science Foundation of China (Grant No. 62473139), in part by the Hunan Provincial Research and Development Project (Grant No. 2025QK3019), in part by the State Key Laboratory of Autonomous Intelligent Unmanned Systems (the opening project number ZZKF2025-2-10), and in part by the Deutsche Forschungsgemeinschaft (DFG, German Research Foundation) - SFB 1574 - 471687386. This research was partially funded by the Ministry of Education and Science of Bulgaria (support for INSAIT, part of the Bulgarian National Roadmap for Research Infrastructure).

%% file: sec/X_suppl.tex
In the supplementary material, we provide a more detailed description of the experimental setup, including all hyperparameters, additional qualitative results, additional quantitative results, and extensive ablation studies, along with the results of ProOOD (+CGFormer). 
We further discuss the limitations of our proposed approach and point out future research directions.

\section{Additional Qualitative Results}
\noindent\textbf{3D Occupancy Prediction Results.}
\begin{figure*}
    \centering
    \includegraphics[width=1\linewidth]{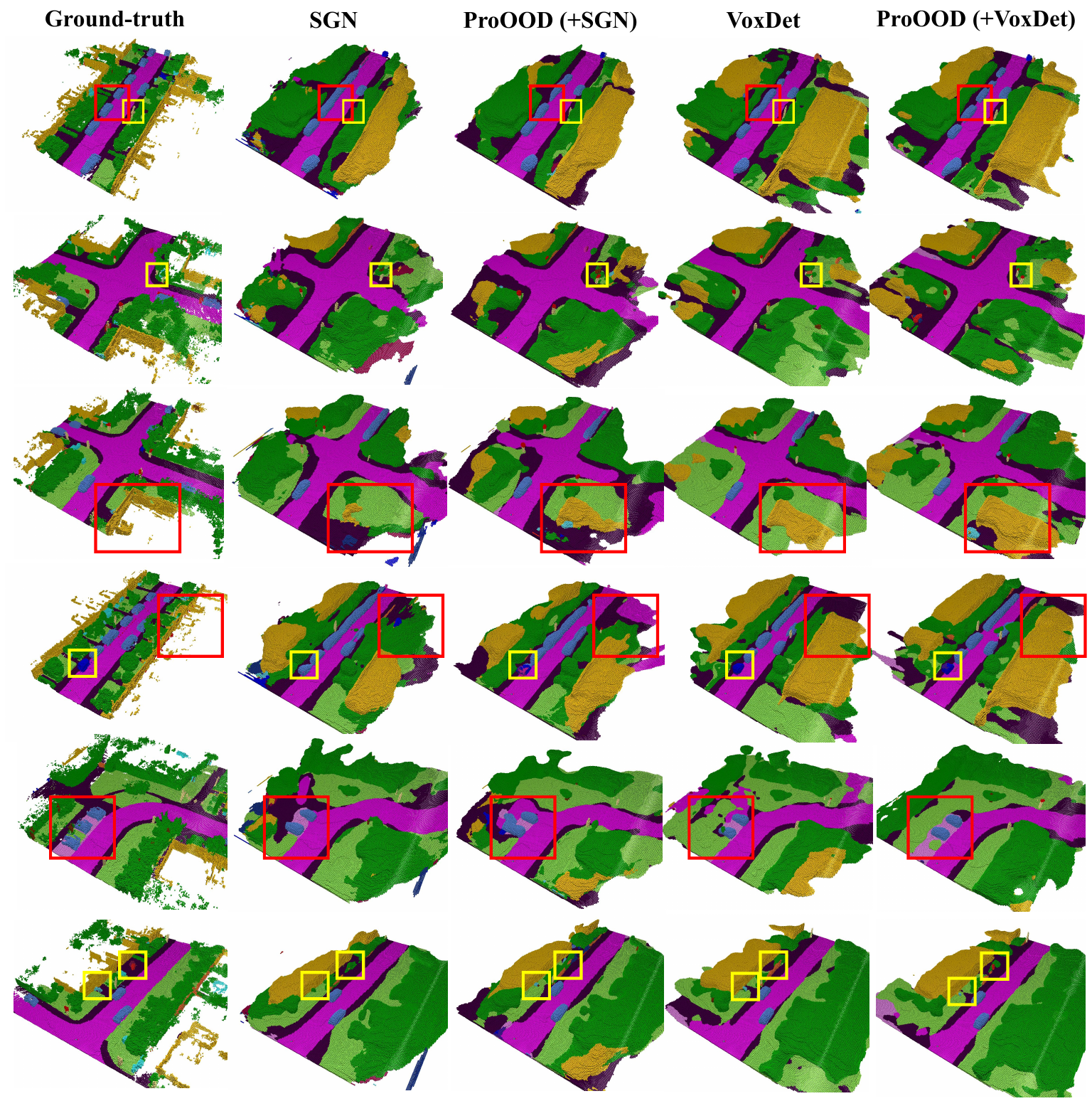}
    \caption{Qualitative results of 3D occupancy prediction on the SemanticKITTI validation set~\cite{behley2019semantickitti}.}
    \label{fig:visual4occ}
\end{figure*}
As shown in Figure~\ref{fig:visual4occ}, we present qualitative comparisons between SGN~\cite{mei2024camera} and VoxDet~\cite{li2025voxdet}, demonstrating the effectiveness of our approach from a voxel prototype-guided perspective. Benefiting from the Prototype-Guided Semantic Imputation module's ability to complete occluded regions, our method improves environmental consistency and achieves more effective instance-level separation. Furthermore, as indicated by the yellow boxes, the enhanced model leveraging Prototype-Guided Tail Mining significantly improves the recognition performance for long-tail categories.

\noindent\textbf{Out-of-Distribution Detection Results.}
\begin{figure*}
    \centering
    \includegraphics[width=1\linewidth]{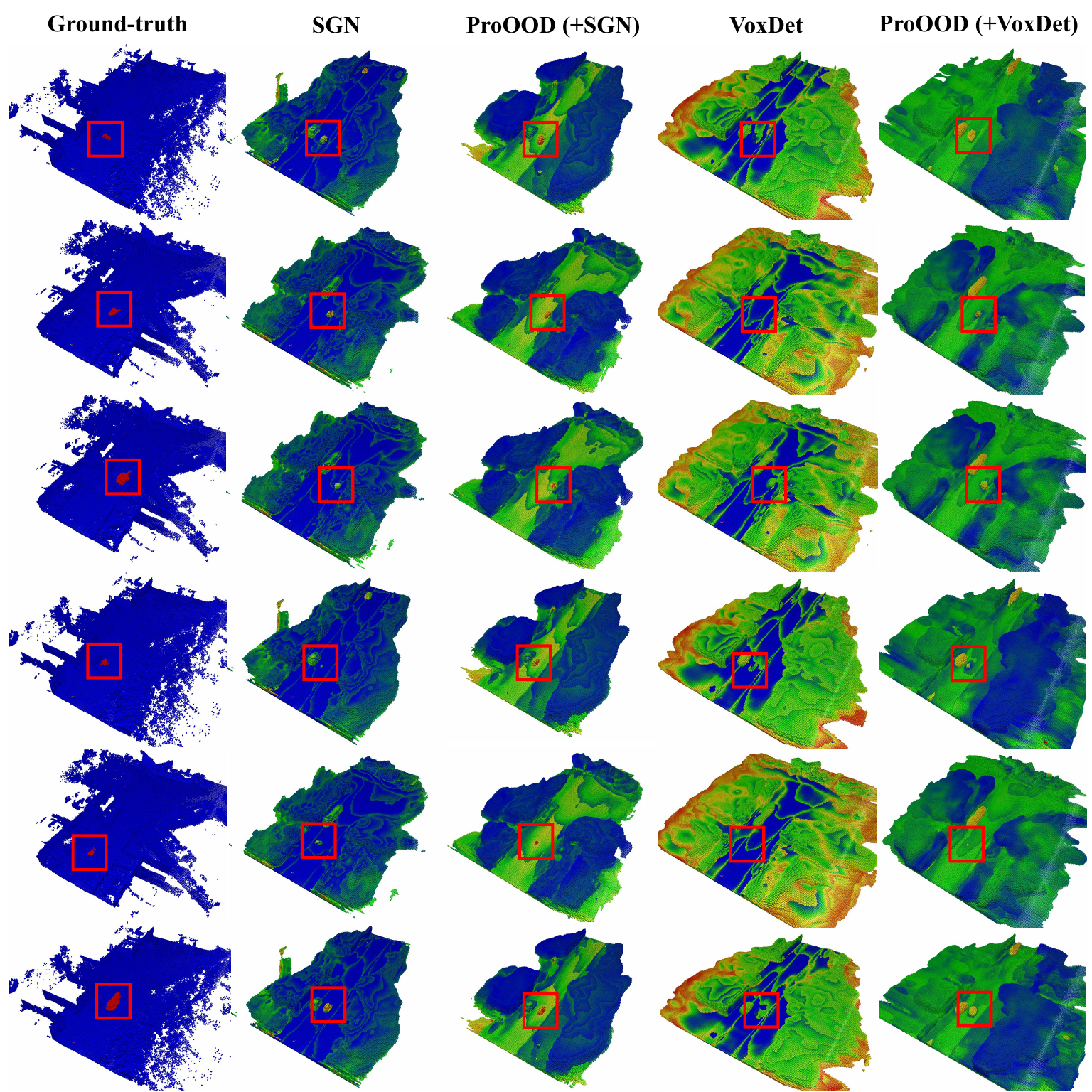}
    \caption{Qualitative results of out-of-distribution detection on the 07VAA sequence of VAA-KITTI ~\cite{zhang2025occood}.}
    \label{fig:visual4ood}
\end{figure*}
As illustrated in Figure~\ref{fig:visual4ood}, our method demonstrates superior performance in detecting OOD voxels. In comparison to the baseline method SGN~\cite{mei2024camera}, ProOOD (+SGN) achieves higher accuracy in identifying OOD voxels. It is worth noting that the highlighted regions in the figure correspond to voxels of the "road" class, a result attributable to the conservative strategy employed. In contrast, the baseline method VoxDet~\cite{li2025voxdet} exhibits almost no OOD detection capability. Although VoxDet achieves stronger in-distribution accuracy, its overconfidence leads to a failure in detecting OOD voxels. ProOOD (+VoxDet) effectively mitigates this overconfidence tendency, thereby endowing the model with a certain level of OOD detection capability.

\section{Additional Quantitative Results}
\input{table/semantickitti_val}
Table~\ref{tab:kitti_val} presents the comparative results on the SemanticKITTI validation set, including the performance of ProOOD (+CGFormer). 
Our ProOOD framework consistently outperforms all respective baselines, achieving up to $18.86$ mIoU and $8.46$ tail mIoU, demonstrating a significant improvement in long-tail modeling capability. Specifically, ProOOD (+SGN) improves SGN~\cite{mei2024camera}'s tail mIoU by $0.82$, while ProOOD (+VoxDet) enhances VoxDet~\cite{li2025voxdet}'s tail mIoU by $1.02$. 
Furthermore, we validate our approach on a state-of-the-art method, CGFormer~\cite{yu2024context}, where it also brings consistent gains of $0.60$ mIoU and $0.44$ tail mIoU. With the exception of the truck class, which exhibits performance fluctuations due to long-tail effects, our implementation essentially reproduces the results reported in the original paper. This demonstrates both the plug-and-play flexibility and effectiveness of our method.

\noindent\textbf{ECE analysis.}
As in Table~\ref{tab:ece}, on VAA-KITTI, where OOD samples severely worsen calibration for both semantic and tail classes, our method consistently reduces $ECE_{sem}$, $ECE_{geo}$, and $ECE_{tail}$, demonstrating effective mitigation of overconfidence, especially for tail classes, and providing direct quantitative support for our claim.
\input{table/ece}

\section{Additional Ablation Studies}
\input{table/ablation_ema}
\noindent\textbf{Exponential Moving Average (EMA) Momentum.}
Table~\ref{tab:ablation_ema} presents an analysis of the impact of the EMA momentum $\beta$, used for updating global voxel prototypes, on model performance. When $\beta$ is set to $0.01$, both mIoU and tail mIoU are suboptimal, as the prototypes evolve too slowly and remain overly influenced by their initial states. A $\beta$ value of $0.1$ yields the best overall mIoU, though tail mIoU begins to decline. This decline becomes more pronounced at $\beta=0.2$, where excessive update rates lead to prototype instability. We therefore select $\beta=0.05$ as it achieves an optimal balance between mIoU and tail mIoU.

\noindent\textbf{Ablation of Hyperparameters}  
\input{table/ablation_hp}
As in Table~\ref{tab:ablation-delta-eta}, we systematically evaluate the impact of the similarity threshold $\eta$ and margin $\delta$ on both overall and tail performance.  
Our default setting (Exp. 3) achieves the best trade-off, yielding the highest tail-mIoU while maintaining strong mIoU. 
With $\eta=0.1$ (Exp. 1), the relaxed similarity criterion risks mismatched injections, but $\delta=0.1$ filters many ambiguous cases, limiting tail-mIoU degradation to just $0.29$. 
In contrast, reducing $\delta$ to $0.05$ (Exp. 2) weakens inter-class separation, severely harming tail performance. 
Increasing $\delta$ to $0.20$ (Exp. 4) overly excludes difficult voxels, especially in tail classes, lowering tail-mIoU to $4.70$ despite a slight mIoU gain, directly reflecting the reviewer’s concern. Finally, raising $\eta$ to $0.5$ (Exp. 5) restricts guidance to only highly confident voxels, excluding informative but uncertain tail samples and reducing tail-mIoU to $4.38$.

\noindent\textbf{Prototype Warm-up Iterations.}
\input{table/ablation_warmup}
Table~\ref{tab:ablation_iter} presents the impact of prototype warm-up iterations on model performance. When the warm-up iterations $t_{\mathrm{warm}}$ are set to $500$, the tail mIoU remains suboptimal, primarily because the feature representations learned in early stages are insufficient for effective prototype initialization. Setting $t_{\mathrm{warm}}$ to $750$ yields the best tail mIoU, while further increasing it to $1000$ or $1500$ leads to a slight degradation in performance. This suggests that involving prototypes earlier in the training process is beneficial for modeling tail categories, as it helps prevent the accumulation of erroneous features during initial training stages. Based on these observations, we select $t_{\mathrm{warm}} = 750$ as the optimal configuration.

\section{Implementation Details}
\noindent\textbf{Backbone Network.}
Our setup follows the baseline configurations. For ProOOD (+SGN) and ProOOD (+VoxDet), we employ ResNet-50~\cite{he2016deep} as the 2D image feature extraction backbone, while ProOOD (+CGFormer) adopts EfficientNet-B7~\cite{tan2019efficientnet}. 
As for the depth estimation network, we adhere to the unified SSC setup commonly used in recent years, utilizing MobileStereoNet~\cite{shamsafar2022mobilestereonet}.

\noindent\textbf{View Transformation.}  
Our setup follows the baseline configurations. ProOOD (+SGN) adopts the FLoSP from MonoScene~\cite{cao2022monoscene}, which constructs 3D features by sampling 2D features via a 3D-2D projection mapping using camera parameters. This approach projects voxel points to image pixels via geometric transformation, then aggregates multi-view 2D features through averaging while masking out non-visible regions.
In contrast, ProOOD (+CGFormer) and ProOOD (+VoxDet) follow CGFormer~\cite{yu2024context}: given an image $I$, we extract 2D feature $\mathbf{F}^{2D}$ and depth map $Z$, then use a depth refinement module with LSS~\cite{philion2020lift} to estimate depth distribution and generate voxel queries $\mathbf{V_Q} \in \mathbb{R}^{X \times Y \times Z \times C}$ with spatial resolution $128 \times 128 \times 16$ and $C=128$. Each pixel $(u, v)$ is transformed to a 3D point $(x, y, z)$ via camera intrinsic and extrinsic matrices, forming query proposals $\mathbf{Q}$ in voxel space, followed by 3D deformable cross-attention to transfer 2D features into 3D space.

\noindent\textbf{Hyperparameter.}
We align the warm-up iterations $t_{\mathrm{warm}}$ for prototype initialization with the baseline settings. Specifically, $t_{\mathrm{warm}}$ is set to $750$ for ProOOD (+SGN) on both datasets, while for ProOOD (+CGFormer) and ProOOD (+VoxDet), it is set to $1500$ on SemanticKITTI~\cite{behley2019semantickitti} and $1600$ on SSCBench-KITTI-360~\cite{li2024sscbench}. 
The exponential moving average (EMA) momentum $\beta$ for prototype updates is set to $0.05$. 
The quality score thresholds $\theta_{\min}$ and $\theta_{\max}$ are set to $0.3$ and $0.7$, respectively. For Prototype-Guided Semantic Imputation (PGSI), the auxiliary occupancy head is weighted by a loss factor of $0.2$, and the residual blending coefficient $\alpha_{\mathrm{pgsi}}$ is initialized as a learnable parameter with value $0.2$. 
In Prototype-Guided Tail Mining (PGTM), the margin $\eta$, the scaling factor $\delta$, and the top-$K$ selection ratio $K_{\mathrm{top}}$ are set to $0.3$, $0.1$, and $0.02$ for SemanticKITTI and $0.0078$ for SSCBench-KITTI-360, corresponding to the proportion of tail classes in each dataset. 
The loss weights for $\mathcal{L}_{\text{tail}}$ and $\mathcal{L}_{\text{proto}}$ are both set to $1$.

\noindent\textbf{Training Setup.}
We follow the training configurations of each baseline method. 
For ProOOD (+SGN), we use $4$ NVIDIA RTX 3090 GPUs with a batch size of $4$, optimized with AdamW~\cite{loshchilov2017decoupled} at an initial learning rate of \(2 \times 10^{-4}\) and weight decay of \(1 \times 10^{-2}\). 
For ProOOD (+CGFormer), we employ $4$ NVIDIA RTX 3090 GPUs with a batch size of $4$, utilizing a cosine annealing schedule with $5\%$ warm-up, a maximum learning rate of \(3 \times 10^{-4}\), weight decay of $0.01$, and Adam parameters \(\beta_1 = 0.9\), \(\beta_2 = 0.99\). 
For ProOOD (+VoxDet), we similarly adopt a cosine annealing schedule with $5\%$ warm-up, maximum learning rate of \(3 \times 10^{-4}\), weight decay of $0.01$, and \(\beta_1 = 0.9\), \(\beta_2 = 0.99\), but adjust the batch size per GPU to accommodate memory constraints, deviating from the original VoxDet~\cite{li2025voxdet} configuration which used $2$ A100 GPUs.

\noindent\textbf{3D Occupancy Prediction Evaluation Configuration.}
We evaluate ProOOD on SemanticKITTI, SSCBench-KITTI-360, VAA-KITTI, and VAA-KITTI-360 for 3D occupancy prediction performance. 
To ensure fair comparison and align with the evaluation protocols of the latter three datasets that only provide single-frame benchmarks without temporal assessment, all models are evaluated without utilizing temporal information. Since the latter three datasets do not provide stereo images, we follow OccOoD~\cite{zhang2025occood} and replace the stereo depth estimation model with the monocular depth model SPI~\cite{lavreniuk2024spidepthstrengthenedposeinformation}.

\noindent\textbf{Out-of-Distribution Detection Evaluation Configuration.}
We evaluate ProOOD on VAA-KITTI, VAA-KITTI-360, and VAA-STU datasets for unsupervised out-of-distribution detection. Specifically, as VAA-KITTI is derived from sequences 07, 09, and 10 of SemanticKITTI, we train our model on sequences 00--06 of SemanticKITTI using the aforementioned monocular depth estimation approach, then evaluate OOD detection on VAA-KITTI. Similarly, for SSCBench-KITTI-360, we employ the same monocular depth model with its default training sequences and assess the OOD performance on VAA-KITTI-360. For VAA-STU, OOD testing is performed directly using weights pre-trained on SemanticKITTI.

\noindent\textbf{Tail classes.}
Tail classes are defined as those with low occurrence frequency in the dataset.
In SemanticKITTI, we classify a class as long-tail if it constitutes less than $0.60\%$ of the total points. 
As summarized in Table~\ref{tab:semantic_kitti_classes}, this includes $11$ classes: \texttt{other-ground}, \texttt{truck}, \texttt{bicycle}, \texttt{motorcycle}, \texttt{other-vehicle}, \texttt{trunk}, \texttt{person}, \texttt{bicyclist}, \texttt{motorcyclist}, \texttt{pole}, and \texttt{traffic-sign}.
Similarly, in SSCBench-KITTI-360, long-tail classes are defined as those representing fewer than $0.30\%$ of total points. 
As shown in Table~\ref{tab:sscbench_kitti360_classes}, there are $7$ such classes: \texttt{bicycle}, \texttt{motorcycle}, \texttt{truck}, \texttt{person}, \texttt{pole}, \texttt{traffic-sign}, and \texttt{other-object}.

\begin{table}[t!]
\centering
\begin{tabular}{l c c}
\hline
\textbf{Class} & \textbf{Percentage} & \textbf{Tail} \\
\hline
road          & 15.30\% & -- \\
sidewalk      & 11.13\% & -- \\
parking       & 1.12\%  & -- \\
other-ground  & 0.56\%  & \checkmark \\
building      & 14.10\% & -- \\
car           & 3.92\%  & -- \\
truck         & 0.16\%  & \checkmark \\
bicycle       & 0.03\%  & \checkmark \\
motorcycle    & 0.03\%  & \checkmark \\
other-vehicle & 0.20\%  & \checkmark \\
vegetation    & 39.30\% & -- \\
trunk         & 0.51\%  & \checkmark \\
terrain       & 9.17\%  & -- \\
person        & 0.07\%  & \checkmark \\
bicyclist     & 0.07\%  & \checkmark \\
motorcyclist  & 0.05\%  & \checkmark \\
fence         & 3.90\%  & -- \\
pole          & 0.29\%  & \checkmark \\
traffic-sign  & 0.08\%  & \checkmark \\
\hline
\end{tabular}
\caption{Class distribution in SemanticKITTI~\cite{behley2019semantickitti}. 
Tail classes (proportion $< 0.60\%$) are marked with \checkmark.}
\label{tab:semantic_kitti_classes}
\end{table}

\begin{table}[h]
\centering
\begin{tabular}{l c c}
\hline
\textbf{Class} & \textbf{Percentage} & \textbf{Tail} \\
\hline
car             & 2.85\%  & -- \\
bicycle         & 0.01\%  & \checkmark \\
motorcycle      & 0.01\%  & \checkmark \\
truck           & 0.16\%  & \checkmark \\
other-vehicle   & 5.75\%  & -- \\
person          & 0.02\%  & \checkmark \\
road            & 14.98\% & -- \\
parking         & 2.31\%  & -- \\
sidewalk        & 6.43\%  & -- \\
other-ground    & 2.05\%  & -- \\
building        & 15.67\% & -- \\
fence           & 0.96\%  & -- \\
vegetation      & 41.99\% & -- \\
terrain         & 7.10\%  & -- \\
pole            & 0.22\%  & \checkmark \\
traffic-sign    & 0.06\%  & \checkmark \\
other-structure & 4.33\%  & -- \\
other-object    & 0.28\%  & \checkmark \\
\hline
\end{tabular}
\caption{Class distribution in SSCBench-KITTI-360. 
Tail classes (proportion $< 0.30\%$) are marked with \checkmark.}\
\label{tab:sscbench_kitti360_classes}
\end{table}

\begin{table}[]
    \centering
    \resizebox{0.4\textwidth}{!}{
    \begin{tabular}{@{}lcc@{}}
        \toprule
        Methods & mIoU & \emph{Tail} mIoU \\
        \midrule
        VoxFormer & 12.86 & 2.87 \\
        Ours (+VoxFormer) & \textbf{13.32} \textcolor{codegreen}{(+0.46)} & \textbf{3.15} \textcolor{codegreen}{(+0.28)} \\
        CGFormer & 15.64 & 4.18 \\
        Ours (+CGFormer) & \textbf{16.24} \textcolor{codegreen}{(+0.60)} & \textbf{4.62} \textcolor{codegreen}{(+0.44)} \\
        \bottomrule
    \end{tabular}}
    \caption{Plug-and-play performance.}
    \label{tab:plug-and-play}
\end{table}
\input{table/model_efficiency}
\input{table/fps}
\section{Plug-and-play Effectiveness}
We further apply our method to the previous state-of-the-art CGFormer~\cite{yu2024context} and the widely adopted VoxFormer~\cite{li2023voxformer}. As shown in Table~\ref{tab:plug-and-play}, consistent performance gains on both models further demonstrate its plug-and-play capability.

\begin{figure*}[!t]
    \centering
    \includegraphics[width=0.75\linewidth]{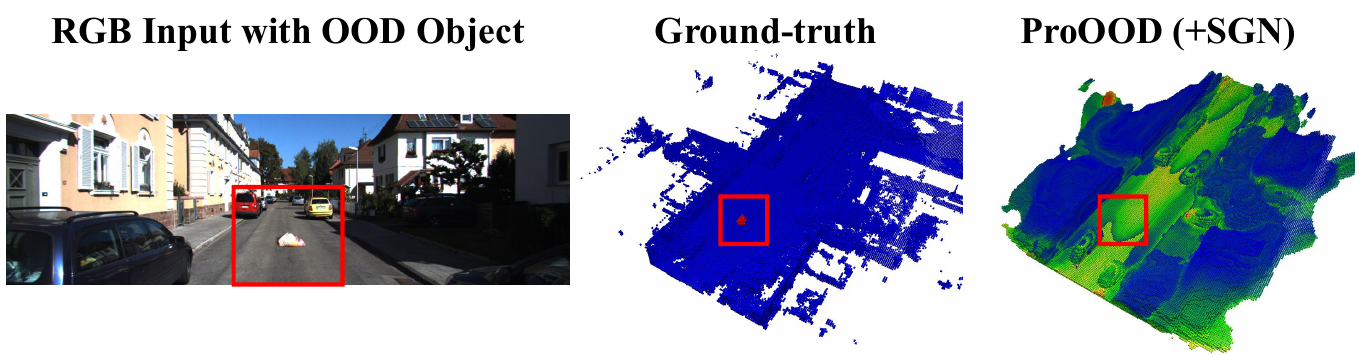}
    \caption{Failure case due to inaccurate occupancy prediction for OOD voxels.}
    \label{fig:failure}
\end{figure*}

\begin{figure*}[!t]
    \centering
    \includegraphics[width=0.6\linewidth]{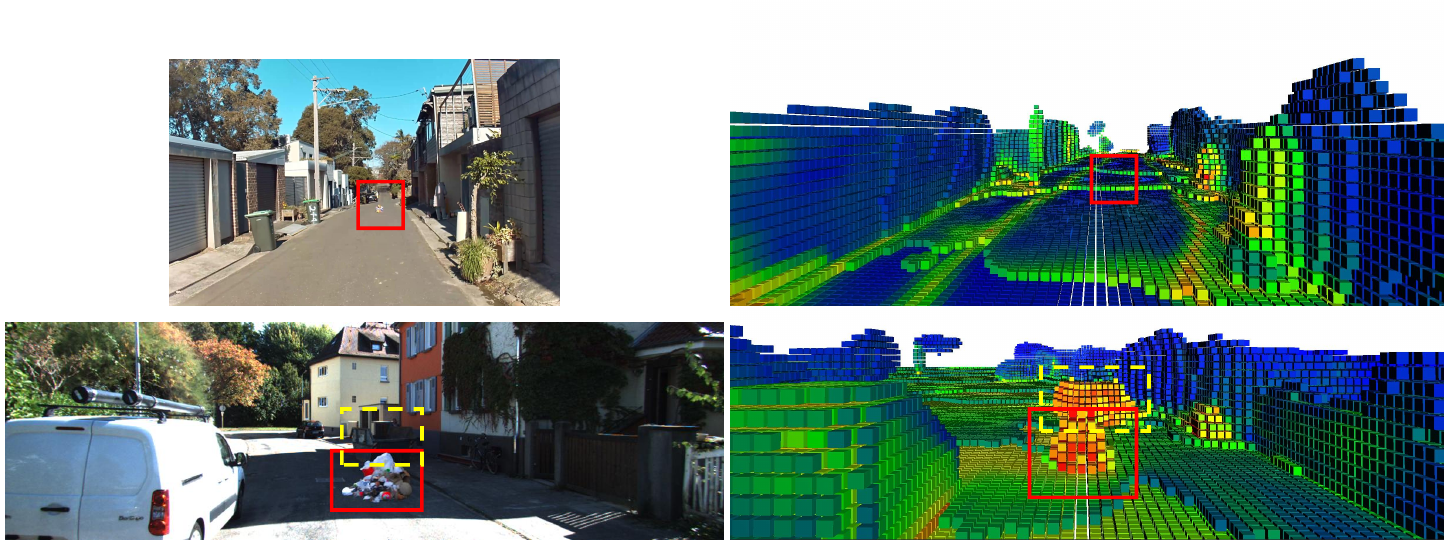}
    \caption{Failure cases under challenging long-range and heavy occlusion scenarios.}
    \label{fig:farrange}
\end{figure*}

\section{Model Efficiency}
As shown in Table~\ref{tab:efficiency}, our proposed method introduces minimal computational overhead while significantly enhancing out-of-distribution detection capabilities. 
For the SGN~\cite{mei2024camera} baseline, we observe a parameter increase of only $0.28M$, where the core prototype learning components contribute merely $0.01M$ parameters, with the remaining overhead attributed to an auxiliary occupancy prediction head. 
More notably, when integrated with VoxDet~\cite{li2025voxdet}, our framework adds only $0.01M$ parameters and $0.40G$ FLOPs, demonstrating exceptional efficiency. This negligible cost stems from our strategic reuse of VoxDet's existing auxiliary prediction modules, eliminating the need for additional architectural components.
These results underscore the computational efficiency of our approach, proving that substantial improvements in OOD detection can be achieved with minimal impact on model complexity and inference cost.

\noindent\textbf{FPS and GPU memory.}
As in Table~\ref{tab:fps-mem}, our method improves performance over the baseline with negligible FPS or memory overhead, demonstrating its efficiency.

\section{Failure Case Analyses}

As illustrated in Figure~\ref{fig:failure}, a typical failure mode occurs when small OOD objects are not reliably activated in the occupancy volume. Since ProOOD leverages occupancy features for OOD scoring, incorrect occupancy predictions directly lead to missed detections. This interdependence suggests that improving occupancy accuracy, particularly for rare and small categories, is crucial. 

Further limitations manifest under extreme geometric conditions, as shown in Figure~\ref{fig:farrange}. In the first row, a distant OOD object is missed due to its small scale and weak geometric or semantic cues. In the second row, while foreground OOD instances are correctly identified, heavily occluded background regions are erroneously classified as OOD, caused by ambiguous semantic boundaries during 2D-to-3D feature lifting.

Potential remedies include: (1) adopting stronger 2D backbones with higher-resolution features; (2) designing object-aware refinement modules to enhance representation learning for small instances; and (3) designing a specialized loss function that imposes a higher penalty on misclassifying small-object voxels to address their current under-representation in the optimization objective.

\section{Limitations and Future Work}
Although ProOOD achieves state-of-the-art performance on OccOoD, it critically relies on the quality of depth estimation, a dependency stemming from the mainstream framework it extends. Since depth estimation is currently implemented as an external module and not integrated into the end-to-end training pipeline, errors from this component can propagate through the system. 
This modular design also constrains the model’s ability to generalize across diverse scenes. 
A promising direction for future work is to incorporate depth estimation directly into the framework, enabling joint optimization and reducing error propagation to enhance performance and generalization.
\section{Societal Impacts}
This research on 3D occupancy prediction and out-of-distribution detection holds broad potential for societal applications, particularly in domains such as autonomous driving and robotic systems. The improved capability to recognize rare objects and detect unknown categories can enhance the safety of autonomous vehicles in complex urban environments, potentially reducing accident risks caused by unexpected obstacles. However, it should be noted that all perception systems have inherent limitations. The failure of our method to detect small or distant objects may still pose risks in safety-critical scenarios. Our reliance on depth estimation quality and the performance limitations in diverse environments indicate that further optimization is required before real-world deployment. We encourage the research community to not only focus on improving technical metrics but also to strive for enhancing the robustness and reliability of such systems under varying environmental conditions.

%% file: table/semantickitti_val.tex
\begin{table*}[!htb]
    \centering
    \resizebox{\textwidth}{!}{%
        \fontsize{7.6pt}{11pt}\selectfont
        \setlength{\tabcolsep}{2.5pt}
        \begin{tabular}{@{}l|*{19}{c}|c|c@{}}
            \toprule
            Method 
            
            & \rotatebox{90}{{\textcolor{red}{\rule{0.5em}{0.5em}}}\hspace{0.3em}road \scalebox{0.7}{(15.30\%)}} 
            & \rotatebox{90}{{\textcolor{pink}{\rule{0.5em}{0.5em}}}\hspace{0.3em}sidewalk \scalebox{0.7} {(11.13\%)}} 
            & \rotatebox{90}{{\textcolor{orange}{\rule{0.5em}{0.5em}}}\hspace{0.3em}parking \scalebox{0.7}{(1.12\%)}} 
            & \rotatebox{90}{{\textcolor{yellow}{\rule{0.5em}{0.5em}}}\hspace{0.3em}other-grnd. \scalebox{0.7}{(0.56\%)}} 
            & \rotatebox{90}{{\textcolor{green}{\rule{0.5em}{0.5em}}}\hspace{0.3em}building \scalebox{0.7}{(14.1\%)}} 
            & \rotatebox{90}{{\textcolor{blue}{\rule{0.5em}{0.5em}}}\hspace{0.3em}car \scalebox{0.7}{(3.92\%)}} 
            & \rotatebox{90}{{\textcolor{purple}{\rule{0.5em}{0.5em}}}\hspace{0.3em}truck \scalebox{0.7}{(0.16\%)}} 
            & \rotatebox{90}{{\textcolor{brown}{\rule{0.5em}{0.5em}}}\hspace{0.3em}bicycle \scalebox{0.7}{(0.03\%)}} 
            & \rotatebox{90}{{\textcolor{cyan}{\rule{0.5em}{0.5em}}}\hspace{0.3em}motorcycle \scalebox{0.7}{(0.03\%)}} 
            & \rotatebox{90}{{\textcolor{magenta}{\rule{0.5em}{0.5em}}}\hspace{0.3em}other-veh. \scalebox{0.7}{(0.20\%)}} 
            & \rotatebox{90}{{\textcolor{lime}{\rule{0.5em}{0.5em}}}\hspace{0.3em}vegetation \scalebox{0.7}{(39.3\%)}} 
            & \rotatebox{90}{{\textcolor{teal}{\rule{0.5em}{0.5em}}}\hspace{0.3em}trunk \scalebox{0.7}{(0.51\%)}} 
            & \rotatebox{90}{{\textcolor{violet}{\rule{0.5em}{0.5em}}}\hspace{0.3em}terrain \scalebox{0.7}{(9.17\%)}} 
            & \rotatebox{90}{{\textcolor{gray}{\rule{0.5em}{0.5em}}}\hspace{0.3em}person \scalebox{0.7}{(0.07\%)}} 
            & \rotatebox{90}{{\textcolor{black}{\rule{0.5em}{0.5em}}}\hspace{0.3em}bicyclist \scalebox{0.7}{(0.07\%)}} 
            & \rotatebox{90}{{\textcolor{lightgray}{\rule{0.5em}{0.5em}}}\hspace{0.3em}motorcyclist \scalebox{0.7}{(0.05\%)}} 
            & \rotatebox{90}{{\textcolor{olive}{\rule{0.5em}{0.5em}}}\hspace{0.3em}fence \scalebox{0.7}{(3.90\%)}}
            & \rotatebox{90}{{\textcolor{navy}{\rule{0.5em}{0.5em}}}\hspace{0.3em}pole \scalebox{0.7}{(0.29\%)}} 
            & \rotatebox{90}{{\textcolor{maroon}{\rule{0.5em}{0.5em}}}\hspace{0.3em}traf.-sign \scalebox{0.7}{(0.08\%)}} 
            & mIoU & \shortstack{\emph{Tail} \\ mIoU}\\
            \midrule 
            \multicolumn{22}{c}{\textbf{\emph{SemanticKITTI}}} \\
            \midrule
            \hline
            
            MonoScene~\cite{cao2022monoscene}  
            
             & 57.47 & 27.05 & 15.72 
            & 0.87 & 14.24 & 23.55 & 7.83 & 0.20 & 0.77 & 3.59
            & 18.12 & 2.57 & 30.76 & 1.79 & 1.03 & 0.00 
            & 6.39 & 4.11 & 2.48 & 11.50 &2.29\\
            
            TPVFormer~\cite{huang2023tri}  
            
             & 56.50 & 25.87 & 20.60 
            & 0.85 & 13.88 & 23.81 & 8.08 & 0.36 & 0.05 & 4.35
            & 16.92 & 2.26 & 30.38 & 0.51 & 0.89 & 0.00 
            & 5.94 & 3.14 & 1.52 & 11.36 &2.00\\
            
            VoxFormer~\cite{li2023voxformer}  
            
             & 54.76 & 26.35 & 15.50 
            & 0.70 & 17.65 & 25.79 & 5.63 & 0.59 & 0.51 & 3.77
            & 24.39 & 5.08 & 29.96 & 1.78 & 3.32 & 0.00 
            & 7.64 & 7.11 & 4.18 & 12.35 &2.97\\
            
            OccFormer~\cite{zhang2023occformer}  
            
             & 58.85 & 26.88 & 19.61 
            & 0.31 & 14.40 & 25.09 & 25.53 & 0.81 & 1.19 & 8.52
            & 19.63 & 3.93 & 32.62 & 2.78 & 2.82 & 0.00 
            & 5.61 & 4.26 & 2.86 & 13.46 &4.82\\
            
            NDC-scene~\cite{yao2023ndc}  
            
             & 59.20 & 28.24 & 21.42 
            & 1.67 & 14.94 & 26.26 & 14.75 & \textbf{1.67} & 2.37 & 7.73
            & 19.09 & 3.51 & 31.04 & 3.60 & 2.74 & 0.00 
            & 6.65 & 4.53 & 2.73 & 12.70& 4.12\\
            HASSC~\cite{wang2024not}  
             &57.05 &28.25 &15.90 &1.04 &19.05 &27.23 &9.91 &0.92 &0.86 &5.61 &25.48 &6.15 &32.94 &2.80 &4.71 &0.00 &6.58 &7.68 &4.05 &13.48&3.97\\     
            
            Symphonies~\cite{jiang2024symphonize}
            
             & 57.11 & 27.79 &  16.97
            & 1.22& 21.74 & 29.08 & 14.36 &3.75 & 3.06 & 14.38
            & 25.32 & 6.87 & 29.70 & \textbf{4.27} & 2.76 & \textbf{0.08}
            & 8.71 & 9.50 & 6.24 & 14.89& 6.04\\
            \midrule   
            SGN~\cite{mei2024camera}
              &59.32 &\textbf{30.51} &18.46 &\textbf{0.42} &21.43 &\textbf{31.88} &13.18 &0.58 &0.17 &\textbf{5.68} &25.98 &7.43 &\textbf{34.42} &1.28 &1.49 &0.00 &9.66 &9.83 &4.71 &14.55&4.07\\
            
            \rowcolor{mygray}ProOOD (+SGN)
             & \textbf{59.35} & 27.94 & \textbf{20.32} & 0.08 &\textbf{21.60}& 31.50 & \textbf{14.54} & \textbf{0.82} & \textbf{0.91} & 5.67 & \textbf{26.06} & \textbf{8.70} & 34.05 & \textbf{3.01} & \textbf{2.33} & 0.00 & \textbf{9.83} & \textbf{10.88} & \textbf{6.85} & \textbf{14.97} & \textbf{4.89} \\

            \midrule   
            CGFormer*~\cite{yu2024context} 
            & 64.79&32.11 & 21.80&0.02 & \textbf{23.19}&\textbf{33.94} &\textbf{1.50} & 2.90& \textbf{4.14}& \textbf{7.21}& \textbf{27.24}&7.56 &38.16 & 1.75& 3.06&0.00 & 9.90& \textbf{10.71}& 7.12& 15.64&4.18\\    \rowcolor{mygray}ProOOD (+CGFormer)        
            & \textbf{66.87} &\textbf{33.61} &\textbf{24.63} & \textbf{0.34}& 22.61&33.78 &1.42 & \textbf{5.50}& 3.41&5.94 & 27.01& \textbf{8.18}&\textbf{38.51} &\textbf{2.87} & \textbf{5.33}& 0.00&\textbf{10.70} & 10.62&\textbf{7.22} & \textbf{16.24}&\textbf{4.62}\\

            \midrule   
            VoxDet*~\cite{li2025voxdet} 
              & 66.50 & 34.47& 19.83& 0.36& \textbf{24.77}& \textbf{34.63}&24.74 & \textbf{3.79}& 5.41&13.23 &28.96 & 9.01&\textbf{41.57} &3.57 & 2.89& 0.00& \textbf{10.89}& \textbf{11.71}&7.18 &18.08 &7.44\\
            \rowcolor{mygray}ProOOD (+VoxDet)
             & \textbf{67.18} & \textbf{34.87}& \textbf{23.64}& \textbf{0.59}&24.54 &34.09 &\textbf{30.60} & 3.75& \textbf{5.95}& \textbf{15.57}& \textbf{29.18}& \textbf{9.67}&41.25 & \textbf{3.75}&\textbf{3.95} &0.00 &10.59 & 11.62&\textbf{7.57} & \textbf{18.86}&\textbf{8.46}\\ 
      
            \bottomrule
        \end{tabular}%
    }
    \caption{Quantitative results on the SemanticKITTI val set~\cite{behley2019semantickitti} (single-frame evaluation for fair comparison). * denotes results reproduced using the official code implementation.}
    \label{tab:kitti_val}
\end{table*}

%% file: table/ece.tex
\begin{table}[h!]
    \centering
    \small

    \resizebox{0.48\textwidth}{!}{
    \begin{tabular}{@{}l|ccc|ccc@{}}
    \toprule
    Methods&$ECE_{sem}$$\downarrow$&$ECE_{tail}$$\downarrow$& $ECE_{geo}$$\downarrow$ & mIoU$\uparrow$& \emph{Tail} mIoU$\uparrow$ &AuROC$\uparrow$ \\
    \midrule
    SGN &51.10&57.49&4.93& 11.10& 2.20 & 60.51\\
    Ours (+SGN)&\textbf{47.28}&\textbf{53.91}&\textbf{4.41} & \textbf{11.86}& \textbf{2.76} &\textbf{64.31}\\
    \bottomrule

    \end{tabular}}
        \caption{Quantitative evaluation of calibration quality.}
    \label{tab:ece}
\end{table}

%% file: table/ablation_ema.tex
\begin{table}[h!]
    \centering
    \small
    \begin{tabular}{@{}c|c|c@{}}
    \toprule
    EMA Momentum& mIoU& \emph{Tail} mIoU\\
    \midrule
    0.01 & 14.75& 4.32\\
    0.05 & 14.97& \textbf{4.89}\\
    0.10 & \textbf{15.14}& 4.55\\
    0.20 & 15.01& 4.36\\
     \bottomrule
    \end{tabular}
    \caption{Ablation study on the impact of different EMA momentum parameters for 3D occupancy prediction on the SemanticKITTI validation set~\cite{behley2019semantickitti}.}
    \label{tab:ablation_ema}
\end{table}

%% file: table/ablation_hp.tex
\begin{table}[h!]
    \centering
    \resizebox{0.35\textwidth}{!}{
    \begin{tabular}{c|cc|cc}
        \toprule
        Exp. ID & \textbf{$\eta$}& \textbf{$\delta$}  & mIoU & \emph{Tail} mIoU \\
        \midrule
        1 & 0.1 & 0.10 & 14.89 & 4.60 \\
        2 & 0.3 & 0.05 & 14.75 & 4.25 \\
        3 & 0.3 & 0.10 & 14.97 & \textbf{4.89} \\
        4 & 0.3 & 0.20 & \textbf{15.16} & 4.70 \\
        5 & 0.5 & 0.10 & 14.84 & 4.38 \\
        \bottomrule
    \end{tabular}}
    \caption{Ablation study of hyperparameters.}
    \label{tab:ablation-delta-eta}
\end{table}

%% file: table/ablation_warmup.tex
\begin{table}[!h]
    \centering
    \small
    \begin{tabular}{@{}c|c|c@{}}
    \toprule
    Warm-up Iterations& mIoU& \emph{Tail} mIoU\\
    \midrule
    500 &15.08 &4.68 \\
    750 & 14.97& \textbf{4.89}\\
    1000 &\textbf{15.09} & 4.80\\
    1500 & 15.01& 4.73\\
     \bottomrule
    \end{tabular}
    \caption{Ablation on different prototype warm-up iterations for 3D occupancy prediction on the SemanticKITTI validation set~\cite{behley2019semantickitti}.}
    \label{tab:ablation_iter}
\end{table}

%% file: table/model_efficiency.tex
\begin{table}[]
    \centering
    \small
    \resizebox{0.48\textwidth}{!}{
    \begin{tabular}{@{}c|cc|cc@{}}
    \toprule
    Method& SGN~\cite{mei2024camera}& ProOOD (+SGN)& VoxDet~\cite{li2025voxdet}& ProOOD (+VoxDet)\\
    \midrule
    \#Parameter (M)&28.16 & 28.44& 55.34& 55.35\\
     FLOPs (G) & 522.34& 580.66&1311.10 & 1311.48\\
     \bottomrule
    \end{tabular}}
    \caption{Comparison of model efficiency with baselines.}
    \label{tab:efficiency}
\end{table}

%% file: table/fps.tex
\begin{table}[h!]
    \centering
    \small

    \resizebox{0.4\textwidth}{!}{
    \begin{tabular}{@{}c|cc|cc@{}}
    \toprule
    Methods&FPS&Inf. Mem. & mIoU& \emph{Tail} mIoU\\
    \midrule
    SGN&7.6&5.16 (GB)& 14.55& 4.07\\
    Ours (+SGN)&7.0&5.30 (GB) & \textbf{14.97}& \textbf{4.89}\\
    VoxDet&5.0&4.34 (GB)&18.08&7.44\\
    Ours (+VoxDet)&4.9&4.47 (GB) & \textbf{18.86}& \textbf{8.46}\\
    \bottomrule
    \end{tabular}}
        \caption{FPS and GPU memory comparisons.}
    \label{tab:fps-mem}
\end{table}